\documentclass[11pt,a4paper]{article}

\usepackage[T1]{fontenc}
\usepackage[utf8]{inputenc}
\usepackage[margin=2.5cm]{geometry}
\usepackage{times}

\usepackage{amsmath,amssymb}
\usepackage{graphicx}
\usepackage{booktabs}
\usepackage{tabularray}
\UseTblrLibrary{booktabs}
\IfFileExists{siunitx.sty}{\usepackage{siunitx}}{} 
\usepackage{url}
\usepackage{natbib}
\usepackage[bottom]{footmisc}
\usepackage{tikz}
\usetikzlibrary{positioning,arrows.meta,decorations.pathreplacing,fit,backgrounds}
\usepackage{algorithm}
\usepackage{algpseudocode}
\usepackage{xcolor}
\IfFileExists{fontawesome.sty}{\usepackage{fontawesome}}{%
  \typeout{WARNING: fontawesome.sty missing - icons in Figure 1 will be blank.}%
  \providecommand{\faServer}{}\providecommand{\faShield}{}%
  \providecommand{\faLaptop}{}\providecommand{\faGraduationCap}{}%
  \providecommand{\faLock}{}\providecommand{\faCogs}{}%
  \providecommand{\faTablet}{}\providecommand{\faMobile}{}%
}

\usepackage[hidelinks]{hyperref}

\setcitestyle{authoryear,round}

\graphicspath{{img/}}

\makeatletter
\renewenvironment{table*}{\@float{table}}{\end@float}
\renewenvironment{figure*}{\@float{figure}}{\end@float}
\makeatother

\newcommand{\jid}[1]{}
\newcommand{\jtitle}[1]{}
\newcommand{\issn}[1]{}
\newcommand{\copyrightstatement}[1]{}
\newcommand{\jyear}[1]{}
\newcommand{\category}[1]{}

\newenvironment{frontmatter}{}{}

\newenvironment{abstract-en}{\begin{abstract}}{\end{abstract}}
\newenvironment{keywords}{\par\medskip\noindent\textbf{Keywords: }}{\par\medskip}

\newenvironment{declarations}{\section*{Declarations}}{}
\newcommand{\declsubsec}[1]{\par\medskip\noindent\textbf{#1.}\ \ignorespaces}
\newenvironment{acknowledgements}{\declsubsec{Acknowledgements}}{\par}
\newenvironment{funding}{\declsubsec{Funding}}{\par}
\newenvironment{contributions}{\declsubsec{Authors' contributions}}{\par}
\newenvironment{interests}{\declsubsec{Conflicts of interest}}{\par}
\newenvironment{materials}{\declsubsec{Availability of data and materials}}{\par}
\newenvironment{aiusagedeclaration}{\declsubsec{Artificial intelligence (AI) usage declaration}}{\par}
\newenvironment{furtherinformation}{\declsubsec{Further information}}{\par}

\providecommand{\orcidlink}[1]{\href{https://orcid.org/#1}{\texttt{#1}}}

\hypersetup{
  pdftitle={Privacy-Preserving Personalization in Education: A Federated Recommender System for Student Performance Prediction},
  pdfauthor={Rodrigo Tertulino; Ricardo Almeida}
}

\title{\textbf{Privacy-Preserving Personalization in Education:\\
A Federated Recommender System for Student Performance Prediction}}

\author{
  \textbf{Rodrigo Tertulino}\thanks{Corresponding author.}\ \ \orcidlink{0000-0002-7594-9312} \\
  \texttt{rodrigo.tertulino@ifrn.edu.br}
  \and
  \textbf{Ricardo Almeida}\ \ \orcidlink{0009-0006-3447-7431} \\
  \texttt{ricardo.almeida@ifrn.edu.br}
}

\date{%
  Federal Institute of Education, Science and Technology of Rio Grande do Norte (IFRN)\\
  Software Engineering and Automation Research Laboratory (LaPEA)\\
  Natal, Rio Grande do Norte, Brazil\\[1em]
  \today
}

\begin{document}

\begin{frontmatter}
    \maketitle

    \begin{abstract-en}
        \textit{Introduction.} Adaptive learning platforms personalize instruction by predicting which skills each student has mastered and recommending what to practice next, but the interaction records underlying these predictions constitute personal data, often from minors, protected by laws such as the Brazilian LGPD. This study asks whether federated learning, which trains a model without collecting students' records, can preserve both the quality and the personalization of such recommendations. \textit{Methods.} We propose a federated recommender in which each of 1,159 students is a client that keeps its own interaction history and shares only model updates, and we evaluate it on the ASSISTments tutoring data against an identical centralized recommender and other models (XGBoost and logistic regression), under a leakage-free protocol with comparable tuning and an ablation of the model's personalized components. \textit{Results.} Keeping the data on the students' devices cost only about one hundredth of AUC, and the decision that matters in practice, choosing the next skill for each student, was practically unaffected: the two recommenders selected the same skill first for 89\% of students. Standard federated averaging did not learn the representation of each individual student, because each student contributes too few examples; an aggregation scheme that does learn it did not improve predictions, and a logistic model on the aggregate features lost nothing to federation. How each device trains its local copy of the model mattered more than the choice between FedAvg and FedProx. \textit{Discussion.} We conclude that privacy-preserving personalization is feasible at a small cost: the individual student representation can be removed, which also takes the most identifying parameter out of the exchange, and federated educational models should be evaluated on the recommendations they would make to each student, with safeguards against leakage.
    \end{abstract-en}

    \begin{keywords}
        Federated learning; Personalized learning; Recommender systems; Student performance prediction; Data privacy; Educational data mining
    \end{keywords}

\end{frontmatter}

\section*{Note on version 2}

This version supersedes arXiv:2509.10516v1. Three changes are material to the
results, and all of them lower or qualify what v1 reported.

\begin{enumerate}
  \item \textbf{Feature leakage corrected.} In v1, the student and skill
  aggregate features and the binarized target were computed over the full
  cohort before the train/test split. For a given student-skill pair, the
  student's aggregate therefore included that pair's own interactions, from
  which the label was derived, so the label was partially embedded in its own
  feature. The evaluation protocol in this version computes every statistic on
  the training split only, under a chronological per-student split. All
  reported numbers were regenerated; none of the v1 figures should be cited.

  \item \textbf{Confounded baseline replaced.} v1 compared a centralized
  XGBoost against the federated neural recommender and attributed the gap to
  federation. That comparison conflates the learning paradigm with the model
  architecture. This version adds an architecture-matched centralized
  recommender, which isolates the cost of federation; the architecture effect
  turns out to be negligible, and the cost of federation is about one
  hundredth of AUC.

  \item \textbf{Aggregation strategies tested against a null.} v1 reported
  FedProx with $\mu = 0.5$ as the best configuration. With tuning budgets
  matched across strategies and effect sizes reported, the differences between
  FedAvg and FedProx configurations do not support that claim. The local
  training procedure matters more than the choice between them.
\end{enumerate}

This version also adds an ablation of the personalized components, a
per-student evaluation of the recommendation actually made, and a full
reproducibility package.

\clearpage

\section{Introduction}
\label{sec:introduction}

Personalized instruction is one of the most consistently supported ideas in education. Students tutored individually outperform students taught in conventional classrooms by a wide margin; \citet{bloom1984sigma} framed this observation as the ``two-sigma problem'', and intelligent tutoring systems that adapt to each learner's step-by-step work recover a substantial part of that advantage \citep{vanlehn2011tutoring}. Online learning platforms pursue the same goal at scale: they record every attempt a student makes and use these records to decide what each student should practice next, for instance, by estimating how likely the student is to master a skill and choosing activities accordingly \citep{shemshack2020personalized}.

Recommender systems are the component that turns such estimates into personalized learning paths. In technology-enhanced learning, they suggest exercises, learning objects, and study sequences tailored to each learner’s progress \citep{handbook_recsys_2011, drachsler2015panorama}, and Brazilian research has applied them to learning-object repositories and to preventing dropout in distance education \citep{dias2020recommender, Andrade2025}. Their pedagogical value depends on the quality of the underlying prediction. A recommender that misjudges which skills a student has mastered either spends practice time on what the student already knows or moves the student forward before the prerequisites are in place.

The records that make this personalization possible, however, are personal data, and in education, they deserve particular care. Performance histories, error patterns, and time on task describe learners in detail, many of them children and adolescents. The Brazilian \textit{Lei Geral de Proteção de Dados} (LGPD) requires that the data of children and adolescents be processed in their best interest and, like the European General Data Protection Regulation (GDPR), limits processing to what is necessary for its purpose \citep{LGPD2018, GDPR2016, Nascimento2023-cv}. Research on learning analytics has long identified privacy as a central ethical issue \citep{slade2013ethics, pardo2014ethical, drachsler2016delicate}; students weigh privacy principles when deciding whether to share their data \citep{ifenthaler2016perceptions}; and a recent systematic mapping in this journal examines fairness, accountability, transparency, and ethics in learning analytics \citep{oliveira2025fate}. Centralizing every student's interaction history in order to train a recommender is therefore not a neutral engineering choice.

Federated Learning (FL) offers a different arrangement. The model travels to the data, each participant trains it locally, and a server aggregates only the resulting model updates \citep{mcmahan_fl_2017, li_federated_2020}. In the vocabulary of data protection law, FL is a technical means of data minimization: the raw records never leave the device or institution that produced them. In education, FL has been applied to knowledge tracing \citep{wu2021fdkt}, bias-aware performance prediction \citep{10.1145/3511808.3557108}, and institutional learning analytics \citep{fachola2023fl, vanhaastrecht2024flame}. A federated recommender, however, faces a difficulty of its own. The parameters that personalize its predictions, such as a learned representation of each student, are trained only on that student's device, and it is not known whether federated training actually learns them. Whether privacy can be preserved without giving up personalization, therefore, remains an open question.

Answering it requires a fair comparison, and student performance prediction offers several ways to make the comparison unfair. The most informative features for this task are aggregates of a student's own success, and aggregates computed carelessly contain the label being predicted \citep{kaufman2012leakage, kapoor2023leakage}. Success labels are imbalanced, so a threshold-dependent metric such as the F1-Score can reward a model that predicts success for everyone \citep{pelanek2015metrics}. Centralized and federated models are often compared across different architectures and with unequal tuning effort, although hyperparameters (settings chosen before training, such as the learning rate or the size of the network) alone can reorder student models \citep{sarsa2022dlkt}. Finally, aggregation strategies are often compared on a single run, where round-to-round fluctuation can be mistaken for an effect. Each of these problems distorts the measured cost of privacy.

This article proposes and evaluates a federated recommender system for student performance prediction that preserves privacy by keeping each student's interaction records on that student's own client. On the ASSISTments 2009--2010 skill-builder data \citep{Feng2009-iu}, a public benchmark that allows the comparison to be controlled on a single, well-known dataset, we compare it with a centralized recommender of identical architecture under a protocol that controls the distortions above: every example is built as ``past history, future outcome'' so that no feature contains its own label, every model is tuned with the same framework and validation data, and FedAvg and FedProx \citep{fedprox_2020} are compared over ten paired seeds with equivalence tests. An ablation study then removes the personalized components of the recommender to determine what federated training learns and what it loses. The study extends a preliminary analysis of this task \citep{anonymous2025preliminary} whose conclusions did not survive these controls, and we report where and why they changed.

The following research questions guide the study:

\begin{itemize}
  \item[\textbf{RQ1:}] How much predictive performance does a federated recommender for student performance prediction lose relative to a centralized recommender of the same architecture, when both are evaluated without leakage and tuned comparably?
  \item[\textbf{RQ2:}] Does federated training with one client per student learn the personalized components of the recommender, namely the student and skill embeddings, how do the aggregation rule and the local configuration affect them, and does learning them improve predictions?
  \item[\textbf{RQ3:}] Does FedProx, with a fixed or tuned proximal coefficient $\mu$, change predictive performance or convergence stability relative to FedAvg?
\end{itemize}

The article makes four contributions:

\begin{enumerate}
  \item A \textbf{federated recommender system for privacy-preserving student performance prediction}, with one client per student, evaluated against an architecture-matched centralized recommender, XGBoost (a widely used model based on decision trees), and logistic regression under a leakage-free protocol in which every model is tuned by Optuna \citep{akiba2019optuna}, a tool that searches hyperparameters automatically, on the same validation data, and evaluated both globally and within each student, which is the ranking that a recommender uses to choose the next skill.
  \item \textbf{Analytical and experimental evidence on the personalized components} of such a recommender when each client holds a single student. Dense averaging leaves the student embeddings (the learned profile that the model keeps for each student) untrained, a limitation that federated recommendation avoids by keeping user parameters local \citep{ammad2019fcf}; we quantify it in student modeling, show how weight decay masks it, and show that row-wise aggregation learns these embeddings without improving predictions of later performance, whereas the skill embedding, updated by many clients, preserves the within-student ranking.
  \item A \textbf{paired, multi-seed comparison of FedAvg and FedProx} with equivalence tests and convergence diagnostics, showing that the local training configuration matters more than the aggregation rule.
  \item \textbf{Methodological lessons for evaluating federated student models}: leave-one-out aggregate encoding leaks the label, the F1-Score conceals majority-class collapse, and an untuned local configuration can make federation appear thirteen times more costly than it is.
\end{enumerate}

The remainder of the article is organized as follows. Section~\ref{sec:related} reviews related work, and Section~\ref{sec:background} formalizes the prediction task, the federated algorithms, and the evaluation measures. Section~\ref{sec:methodology} describes the data, protocol, and analysis, and Section~\ref{sec:results} presents the results. Section~\ref{sec:discussion} answers the research questions, Section~\ref{sec:limitations} discusses the limitations of the study, Section~\ref{sec:openscience} describes the available materials, and Section~\ref{sec:conclusion} concludes.

\section{Related Work}
\label{sec:related}

\subsection{Recommender systems for personalized learning}

Recommender systems entered education from e-commerce, but learning imposes requirements that product recommendations do not. A good recommendation must suit the learner's goals and current knowledge, respect pedagogical sequencing, and serve learning rather than momentary preference \citep{handbook_recsys_2011, drachsler2015panorama}. Personalized learning, in this sense, adapts content, pace, and path to the individual learner \citep{shemshack2020personalized}, and the recommender is the mechanism that implements this adaptation in a platform. Studies published in Brazilian venues demonstrate the diversity of these applications, ranging from recommending learning objects based on learners' previous choices \citep{dias2020recommender} to combining educational data mining with active methodologies to prevent dropout in distance education \citep{Andrade2025}. Recommenders that choose the next activity rely on an estimate of what the student can already do, which makes student performance prediction their core component. Neural recommenders typically represent each user and each item, here each student and each skill, by a learned \emph{embedding}. An embedding is a short list of numbers that the model keeps for each student (or skill) and adjusts during training; it works like an individual profile card that summarizes characteristics of that student which the shared features do not capture, for example, a tendency to perform better or worse than the student's overall success rate would suggest on certain kinds of skills. The numbers have no predefined meaning: they start at random values and are shaped by the student's own data. Embeddings therefore allow predictions to depart from what the shared features imply; they are the personalized components of the model and, for the student, they are the only individual parameters.

\subsection{Privacy of student data}

The data that feeds educational recommenders raises ethical questions that the learning analytics community has examined for more than a decade. \citet{slade2013ethics} identifies the location and interpretation of data, informed consent, and the management, classification, and storage of data as central issues, and \citet{pardo2014ethical} proposes principles of transparency, student control over data, security, and accountability. The DELICATE checklist translates these concerns into practical requirements for trusted learning analytics \citep{drachsler2016delicate}, and students' own perceptions of these principles shape their willingness to share data \citep{ifenthaler2016perceptions}. In Brazil, the LGPD gives these principles legal force \citep{LGPD2018, Nascimento2023-cv}, recent work has assessed the privacy of Brazilian open educational data \citep{maretto2023dados}, and a systematic mapping in this journal organizes the literature on fairness, accountability, transparency, and ethics in learning analytics \citep{oliveira2025fate}. Most of these proposals are organizational: consent, transparency, and governance. Federated learning adds a technical measure that complements them, since it avoids collecting raw records in the first place; secure aggregation \citep{bonawitz2017secagg} and differential privacy \citep{abadi2016dpsgd} can further protect the exchanged updates.

\subsection{Student performance prediction}

Predicting whether a student will succeed is among the oldest problems of educational data mining \citep{hellas_slr_2018}. At the level of courses and programs, centralized models trained on academic records identify students at risk of failure or dropout \citep{sales_academic_2016, melo_semisupervised_2023}, a line of work well represented in this journal \citep{Baker2011, pereira_deeplearning_2020, Teodoro2020, carneiro_edm_2022} and consolidated by a recent review of dropout prediction \citep{colpo_slr_2024}. At the level of skills within a tutoring system, knowledge tracing models the evolution of a student's mastery from the sequence of attempts, from Bayesian formulations \citep{corbett1994kt} to recurrent networks \citep{piech2015dkt}. The ASSISTments data used here are a standard benchmark for this second family \citep{Feng2009-iu}, and they illustrate how sensitive reported performance is to data preparation: duplicated rows in an early release inflated the apparent advantage of deep knowledge tracing \citep{xiong2016deeper}, and logistic regression over well-chosen aggregate features remains competitive with deep models on datasets of moderate size \citep{gervet2020kt}. Our task sits between the two levels. We predict a student's success rate on a skill in a future window from aggregate descriptors of past behavior, the kind of signal an adaptive platform uses to decide what to recommend.

\subsection{Federated learning under heterogeneous clients}

FedAvg trains a shared model by averaging locally updated weights \citep{mcmahan_fl_2017}. When client distributions differ, local updates drift apart and averaging converges more slowly or to worse solutions \citep{zhao2018noniid, li_federated_2020}. FedProx adds a proximal term that keeps local weights close to the global model \citep{fedprox_2020}; SCAFFOLD corrects drift with control variates \citep{karimireddy2020scaffold}; FedDyn regularizes local objectives dynamically \citep{acar2021feddyn}. The optimizer used on the client and server also matters: adaptive server optimizers can improve convergence in heterogeneous settings \citep{reddi2021adaptive}, and practical guidance emphasizes that local learning rates and epochs must be tuned jointly with the aggregation rule \citep{wang2021fieldguide}. A setting with one client per student is an extreme case of heterogeneity, since each client holds a handful of examples from a single learner. It also poses a specific challenge for models with per-user parameters, such as a recommender's student embeddings, because each parameter is trained by a single client. Federated recommendation has addressed this difficulty since its first formulation. Federated collaborative filtering keeps each user's latent factors on the user's device and aggregates only the item factors \citep{ammad2019fcf}; later work protects the exchanged item updates with homomorphic encryption \citep{chai2021sfmf}, extends the scheme to explicit feedback \citep{lin2021fedrec}, and accelerates its training \citep{muhammad2020fedfast}. Personalized federated learning follows the same principle by keeping some layers local \citep{arivazhagan2019fedper}, and partially local training generalizes it to arbitrary models \citep{singhal2021fedrecon}. The standard FedAvg and FedProx strategies of general-purpose frameworks such as Flower \citep{beutel2020flower}, however, average every parameter densely, and a student model trained with them inherits this behavior.

\subsection{Federated learning in education}

Federated approaches have been proposed for knowledge tracing across schools \citep{wu2021fdkt}, for reducing bias in performance prediction through personalization \citep{10.1145/3511808.3557108}, and for learning analytics across institutions \citep{fachola2023fl}. \citet{vanhaastrecht2024flame}, in turn, compared federated, centralized, and local learning for educational analytics and found that federated performance was comparable to centralized learning. Together, these studies establish feasibility. Their conclusions about the cost of federation, however, rest on how features, metrics, baselines, and tuning were handled, which is the subject of the present study rather than a criticism of any single work.

\subsection{Evaluating student models}

Leakage, the use during training of information unavailable at prediction time, inflates reported performance across fields \citep{kaufman2012leakage, kapoor2023leakage}. Target statistics such as a student's mean success are a well-known channel: computing them over the full data leaks the test labels, and even leave-one-out versions leak the training label, the problem that motivated ordered target statistics in CatBoost \citep{prokhorenkova2018catboost}. For student models, \citet{pelanek2015metrics} discusses how the choice of metric shapes conclusions, and \citet{sarsa2022dlkt} shows that hyperparameter choices and metrics affect both the ranking and the replicability of knowledge tracing models. For comparisons of learning algorithms, repeated runs, non-parametric omnibus tests, and corrections for multiple comparisons are standard recommendations \citep{demsar2006}, and a claim that two methods are equivalent requires an equivalence test rather than a non-significant difference \citep{lakens2017equivalence}.

\subsection{Research gap}

Two gaps follow from this review. First, the studies of federated learning in education report that federated models approach centralized ones, but they do not examine whether the personalized parameters of a federated model are actually learned when each client holds a single student's data, although federated recommendation and partially local schemes keep such parameters on the client precisely because they are hard to train by averaging \citep{ammad2019fcf, singhal2021fedrecon}. Nor do they evaluate federated models within students, the ordering that a recommender uses. Second, to our knowledge, no study of federated student modeling has combined the evaluation safeguards above: features constrained to precede the label, a centralized reference of the same model class, the same tuning procedure on both sides, and repeated runs with equivalence tests. The present study addresses both gaps for a federated recommender of student performance.

\section{Theoretical Foundations}
\label{sec:background}

This section formalizes the prediction task, the two federated algorithms compared, the way these algorithms treat per-student parameters, and the measures used to evaluate the models.

\subsection{Prediction task and model}
\label{sec:task}

Let $\mathcal{H}_u$ be the chronologically ordered interactions of student $u$, each recording a skill and whether the answer was correct. For a label window $W$ of $\mathcal{H}_u$ and a skill $s$ practiced in it, the target is
\begin{equation}
\label{eq:target}
y_{us} = \mathbb{1}\!\left[\frac{1}{|W_s|}\sum_{i \in W_s} c_i \geq \tau\right],
\end{equation}
where $W_s$ are the interactions of $u$ with $s$ inside $W$, $c_i \in \{0,1\}$ indicates a correct answer and $\tau$ is the mastery threshold. The features $x_{us}$ are aggregates computed only on the interactions that precede $W$ (Section~\ref{sec:examples}). The recommender network scores a pair as
\begin{equation}
\label{eq:model}
\hat{p}_{us} = \sigma\!\left(g_\theta\!\left([\,e_u;\, e_s;\, x_{us}\,]\right)\right),
\end{equation}
where $e_u$ and $e_s$ are learned embeddings of the student and the skill, $g_\theta$ is a multilayer perceptron (a small neural network made of fully connected layers), $\sigma$ is the logistic function, and $[\cdot\,;\cdot]$ denotes concatenation. Concretely, $e_u$ is a vector of 32 numbers (the size selected by tuning) reserved for student $u$, and $e_s$ is a vector of the same size for skill $s$. All student vectors are stored together in an \emph{embedding table} with one row per student, and all skill vectors in a second table with one row per skill; the model reads the row of the student and the row of the skill involved in each prediction. Like the other parameters, the rows start with random values and are adjusted during training: a student's row changes only when the model is trained on that student's own examples, which is why it serves as that student's individual profile. The two embeddings are the personalized components of the recommender: they allow its predictions for a particular student and skill to depart from what the shared aggregate features imply, and $e_u$ is the only parameter specific to that student. Under federation, the student-level features are computed on the student's own client, whereas the skill statistic requires per-skill sums from all clients (Section~\ref{sec:models}).

\subsection{FedAvg and FedProx}

Let $K$ be the number of clients, $n_k$ the number of training examples of client $k$, and $F_k(w)$ its empirical loss. In round $t$ (a \emph{communication round}) the server sends the global weights $w^t$ to a sample $S_t$ of clients; each client runs $E$ local epochs, that is, $E$ passes over its own examples, processed in small groups (mini-batches), starting from $w^t$, and returns the resulting weights $w_k^{t+1}$. FedAvg \citep{mcmahan_fl_2017} then sets
\begin{equation}
\label{eq:fedavg}
w^{t+1} = \sum_{k \in S_t} \frac{n_k}{\sum_{j \in S_t} n_j}\, w_k^{t+1}.
\end{equation}
FedProx \citep{fedprox_2020} keeps this aggregation rule and changes the local objective to
\begin{equation}
\label{eq:fedprox}
\min_{w}\; h_k(w) = F_k(w) + \frac{\mu}{2}\,\lVert w - w^t \rVert^2 ,
\end{equation}
so that $\mu > 0$ penalizes departures from the global model and $\mu = 0$ recovers FedAvg. The proximal coefficient interacts with the local learning rate and the number of local steps: a large $\mu$ limits drift, but it also limits how much each round can learn.

\subsection{Embedding tables under aggregation}
\label{sec:embeddings}

Equation~\eqref{eq:fedavg} treats the embedding tables differently from the shared weights of the perceptron. On client $k$, the loss depends only on the rows of the student and the skills present in that client's data, so every other row $r$ receives a zero gradient, the signal that training uses to decide how to change a parameter. Without weight decay, those rows are returned unchanged, and after aggregation
\begin{equation}
\label{eq:dilution}
e_r^{t+1} = e_r^{t} + \sum_{k \in S_t(r)} \frac{n_k}{\sum_{j \in S_t} n_j}\,\bigl(e_{r,k}^{t+1} - e_r^{t}\bigr),
\end{equation}
where $S_t(r) \subseteq S_t$ is the set of sampled clients whose data involve row $r$ and $e_{r,k}^{t+1}$ is the value of the row returned by client $k$. For a student row, $S_t(r)$ contains at most the student's own client: the row moves only in the rounds in which that student is sampled (about one round in ten here), and then by a fraction of roughly $1/|S_t|$ of the local update. A skill row is updated by every sampled client who practiced the skill, so its dilution is milder but still present. In centralized training, by contrast, every row receives its full gradient in every epoch. Equation~\eqref{eq:dilution} therefore predicts that FedAvg learns the embedding tables much more slowly than the shared weights, and that the rows remain close to their random initialization, which in common implementations draws each row from a standard normal distribution. The perceptron then receives, alongside the informative features, embedding inputs that are largely noise.

Local weight decay changes this picture. Weight decay is a standard regularization technique that slightly shrinks every parameter toward zero at each training step, preventing the model from relying on overly large values. With coupled $L_2$ regularization of strength $\lambda$, every local step adds $\lambda e_r$ to the gradient of every row, including the rows absent from the client. Under stochastic gradient descent (SGD), which moves each parameter against its gradient by a step proportional to the learning rate $\eta$, these rows shrink geometrically by a factor $(1-\eta\lambda)$ per step. Under Adam, a widely used variant that rescales each step by running estimates of the gradient's magnitude, the update is normalized coordinate-wise, so a gradient that consists only of $\lambda e_r$ moves each coordinate by a step of order $\eta$ regardless of $\lambda$ \citep{loshchilov2019adamw}. Weight decay thus erases, round after round, the random initialization that FedAvg cannot replace with learned values; under Adam, however, it does so in large, uniform steps that do not depend on the data.

Federated recommendation avoids the dilution of Equation~\eqref{eq:dilution} by aggregating each embedding row only over the clients that use it \citep{ammad2019fcf, lin2021fedrec}:
\begin{equation}
\label{eq:rowwise}
e_r^{t+1} = \sum_{k \in S_t(r)} \frac{n_k}{\sum_{j \in S_t(r)} n_j}\, e_{r,k}^{t+1} \qquad \text{for } S_t(r) \neq \varnothing,
\end{equation}
while rows that no sampled client uses keep their value $e_r^{t}$. For a student row, Equation~\eqref{eq:rowwise} sets the row to the value computed by that student's own client, which is equivalent to keeping the student embedding on the client; skill rows are averaged over the clients that practiced the skill. Local weight decay on rows absent from a client is discarded. Section~\ref{sec:ablation} tests these predictions and whether learning the embeddings improves predictions.

\subsection{Evaluation measures}
\label{sec:measures}

The ROC AUC equals the probability that a randomly chosen positive pair receives a higher score than a randomly chosen negative pair \citep{fawcett2006roc}. It is invariant under monotone transformations of the scores and does not depend on a decision threshold or the class prior; a value below 0.5 indicates that the ranking is inverted. Balanced accuracy is the mean of the true positive rate (TPR) and the true negative rate (TNR) at a threshold $\gamma$,
\begin{equation}
\label{eq:balacc}
\mathrm{BA}(\gamma) = \tfrac{1}{2}\bigl(\mathrm{TPR}(\gamma) + \mathrm{TNR}(\gamma)\bigr) = \tfrac{1}{2}\bigl(1 + J(\gamma)\bigr),
\end{equation}
where $J$ is Youden's index \citep{youden1950, brodersen2010balanced}; any predictor that assigns the same class to every pair has $\mathrm{BA} = 0.5$. The F1-Score, by contrast, rewards the majority class when it is the positive class. Predicting success for every pair gives precision equal to the positive rate $\pi$ and recall 1, hence
\begin{equation}
\label{eq:f1trivial}
F_1^{\text{all-positive}} = \frac{2\pi}{1+\pi},
\end{equation}
which equals 0.783 for the test positive rate of this study ($\pi = 0.643$). Any F1 near this value must therefore be read against it.

To claim that two strategies are equivalent, we use two one-sided tests (TOST) \citep{lakens2017equivalence}. For a paired mean difference $\Delta$ and a margin $\delta$, TOST rejects $H_0\colon |\Delta| \geq \delta$ when both $H_0^{-}\colon \Delta \leq -\delta$ and $H_0^{+}\colon \Delta \geq \delta$ are rejected by one-sided $t$-tests; its $p$ value is the larger of the two. A non-significant difference test, in contrast, provides no evidence of equivalence.

\section{Methodology}
\label{sec:methodology}

\subsection{Data and cohort}

We use a single dataset by design. The aim of this study is not to compare datasets or to find the best predictor for a particular platform, but to examine what protecting students' data costs a recommender and how it affects personalization. Answering that question requires holding everything else fixed and, on a single dataset, controlling for the factors that usually confound the comparison (Section~\ref{sec:introduction}). We chose the ASSISTments 2009--2010 skill-builder data for four reasons: it is public and de-identified; it is one of the most widely used benchmarks in student modeling \citep{Feng2009-iu, piech2015dkt, xiong2016deeper, gervet2020kt}, which makes our results comparable with previous work; it records each student's attempts individually and in order, which allows each student to act as a federated client and every feature to be computed from the past only; and it tags each problem with the skill it exercises, which is the unit a recommender works with.

We use the ``corrected, collapsed'' version of these data, distributed by the ASSISTments project \citep{Feng2009-iu}, which was released after duplicate rows were identified in the original file \citep{xiong2016deeper}. Each row records a student's first attempt at a problem, the skill it exercises, and whether the answer was correct. We discard rows without a student, skill, answer, or order identifier. So that every client and every skill has enough history to build windows, we keep students with at least 50 interactions and skills with at least 100 interactions. We order each student's interactions by \texttt{order\_id}. Because the official documentation does not describe this field as chronological, we checked it against \texttt{opportunity}, which counts a student's successive attempts on a skill and is ordinal by construction; the two orders agree perfectly (mean Spearman rank correlation $\rho = 1.000$, where 1 means identical orderings, over 20,030 student--skill pairs). Table~\ref{tab:cohort} summarizes the resulting cohort.

\begin{table}[ht]
\centering
\caption{Cohort and example construction after filtering. Data: own elaboration from ASSISTments 2009--2010.}
\label{tab:cohort}
\begin{tblr}{
    colspec = {p{0.58\linewidth} X[r]},
    cells   = {font=\small},
    row{1}  = {font=\small\bfseries},
    }
\toprule
Quantity & Value \\
\midrule
Interactions (raw rows / valid / after filtering) & 346,860 / 283,105 / 234,294 \\
Students (federated clients) / skills & 1,159 / 121 \\
Training examples: fit / validation & 17,669 / 1,907 \\
Test examples & 9,575 \\
Positive rate: fit / validation / test & 0.663 / 0.653 / 0.643 \\
Fit examples per client: min / median / max & 1 / 9 / 94 \\
Test pairs whose skill appears in the student's history & 0.503 \\
Spearman $\rho$ between \texttt{order\_id} and \texttt{opportunity} (mean) & 1.000 (20,030 pairs) \\
\bottomrule
\end{tblr}
\end{table}

\subsection{Leakage-free example construction}
\label{sec:examples}

An example is a (student, skill) pair observed in a \emph{label window} of the student's history. Its target (Equation~\eqref{eq:target}) is the pair's success rate in that window binarized at $\tau = 0.70$, the level at which we regard the skill as mastered. Its features are three aggregates computed on the student's history \emph{before} the window: the student's past success rate (\texttt{user\_mean\_correct}), the logarithm of the number of past interactions (\texttt{user\_interaction\_count}), and the skill's past success rate across all students (\texttt{skill\_mean\_correct}). Figure~\ref{fig:windows} shows how the windows are placed in each student's own timeline. Training examples come from two expanding windows (history 0--40\% predicting 40--60\%, and history 0--60\% predicting 60--80\%); test examples use history 0--80\% to predict the last 20\%. Because the windows are positioned within each student's history rather than calendar dates, the skill statistic for an example is computed over the corresponding history windows for all students (for test examples, the first 80\% of each student's history). It therefore never includes any student's label window, but it may include interactions that occurred later in calendar time than another student's test window; this is not leakage of the example's own label, but it means that the skill statistic reflects the cohort as a whole. Ten percent of the training examples, chosen at random, form a validation set; validation, therefore, shares students and windows with training, while the test window lies later in every student's history.

\begin{figure}[ht]
\centering
\begin{tikzpicture}[x=0.075\linewidth, y=0.5cm, font=\scriptsize]
  \foreach \y/\a/\b/\lab in {2/4/6/{Training, window 1}, 1/6/8/{Training, window 2}, 0/8/10/{Test}} {
    \fill[gray!25] (0,\y) rectangle (\a,\y+0.7);
    \fill[blue!45] (\a,\y) rectangle (\b,\y+0.7);
    \node[anchor=east] at (0,\y+0.35) {\lab};
  }
  \foreach \x/\l in {0/0, 4/40, 6/60, 8/80, 10/100} \node[below] at (\x,0) {\l\%};
  \node[anchor=west] at (0,3.3) {\tikz\fill[gray!25] (0,0) rectangle (0.3,0.2); features};
  \node[anchor=west] at (3,3.3) {\tikz\fill[blue!45] (0,0) rectangle (0.3,0.2); labels};
  \node[below] at (5,-0.6) {position in the student's own interaction history};
\end{tikzpicture}
\caption{Placement of feature and label windows in each student's history. Features never overlap the label window of the same example. Data: own elaboration.}
\label{fig:windows}
\end{figure}

This design replaces the construction used in a preliminary analysis of this task, in which features were first computed over the whole cohort and later over the training pool with leave-one-out subtraction of the example's own interactions. The second variant may seem sufficient, but leave-one-out removes the label from the feature, not its influence: for a fixed student, the leave-one-out mean decreases as the pair's own success increases, and, together with the leave-one-out count, it allows the label to be reconstructed. Applied to the three features, the same XGBoost configuration achieves a validation AUC of 0.928 and a test AUC of 0.719 under leave-one-out encoding, a gap that signals leakage; with the windows shown in Figure~\ref{fig:windows}, the validation and test AUCs are close (Section~\ref{sec:results}).

\subsection{Models}
\label{sec:models}

\textbf{RecommenderNet} implements Equation~\eqref{eq:model}: it embeds the student and the skill, concatenates the two embeddings with the three features, and passes the result through two fully connected layers with ReLU activations and dropout (which randomly silences part of the network during training to reduce overfitting), followed by a single output score (logit) trained with binary cross-entropy, the standard loss function for yes/no outcomes. Its embedding size, layer widths, and dropout are tuned (Section~\ref{sec:tuning}). The embedding tables use dense gradients, weight decay is the coupled $L_2$ penalty of the optimizer (not the decoupled form of AdamW), and in federated training, each client creates a fresh optimizer in every round, so that Adam's moment estimates are reset; these are the conditions under which the analysis of Section~\ref{sec:embeddings} applies.

\textbf{Centralized DNN (centralized recommender).} RecommenderNet trained on the pooled training examples with early stopping: training stops when validation AUC has not improved for 8 consecutive epochs (at most 60 epochs), and the best epoch is kept.

\textbf{Centralized XGBoost} \citep{chen_xgboost_2016} on the same inputs, with the encoding of student and skill identifiers (numeric, categorical, or omitted) treated as a hyperparameter; boosting rounds are chosen by early stopping on validation AUC.

\textbf{Federated recommender.} RecommenderNet with the architecture selected for the centralized DNN, trained by FedAvg or FedProx with one client per student. Each round samples $\max(50, \lfloor 0.1 K \rfloor)$ clients, 115 in this cohort, and training runs for a fixed budget of 1{,}000 rounds, chosen before tuning. The checkpoint, that is, the model as it stood after a given round, with the highest validation AUC is retained. We compare FedAvg, FedProx with $\mu \in \{0.1, 0.5, 1.0\}$, and FedProx with $\mu$ tuned. Its data flow is described in Section~\ref{sec:dataflow}.

\textbf{Row-wise aggregation.} The federated recommender trained with FedAvg, but with the embedding tables aggregated by Equation~\eqref{eq:rowwise}; it is evaluated both with the configuration selected for dense FedAvg, which isolates the aggregation rule, and with its own Optuna search.

\textbf{Logistic regression, centralized and federated.} A logistic regression on the three aggregate features, trained centrally with the training loop of the centralized DNN and federatedly with FedAvg. It tests whether a simple and fully federable model suffices, as logistic models over well-chosen features often do in student modeling \citep{gervet2020kt}.

\textbf{Learning-free references.} Predicting success for every pair, a stratified random guess, and ranking pairs by the student's or the skill's past success rate alone.

\textbf{Preliminary configuration.} The preliminary analysis \citep{anonymous2025preliminary} compared a centralized XGBoost with a federated network. To connect with it, we also train RecommenderNet with the original settings of that network (embedding size 10, layers 32 and 16, no dropout, Adam with learning rate $10^{-3}$, batch size 32, and, when federated, five local epochs), both centrally and with FedAvg.

\subsection{Proposed federated recommender and data flow}
\label{sec:dataflow}

Figure~\ref{fig:system} summarizes the proposed system from the perspective of data protection. Each student's device keeps the raw interaction history, the correctness of each answer, and the labels derived from it; it computes the student's aggregate features locally and trains the recommender on them. In each round, the server sends the current global model to the sampled devices and receives only their updated weights, which it combines with Equation~\eqref{eq:fedavg}. The only other quantities that leave the devices are the per-skill counts of attempts and successes required by the skill statistic, which a deployment would obtain through secure aggregation \citep{bonawitz2017secagg}, so that the server learns only the totals over all students. The global model contains the student embedding table, and a given student's row changes only when that student participates; this row is therefore the most identifying element of the exchange, a point we return to in Section~\ref{sec:platforms}.

We assume an honest-but-curious server: the party that trains and aggregates the model follows the protocol but may attempt to infer information about individual students from the data it receives. This fits settings in which the organization that trains the recommender does not hold the records, such as an analytics provider serving several schools, or in which the learning environment runs on the student's device or on the school's own infrastructure. In platforms that grade answers on a central server, as ASSISTments itself does, the server already holds the records, and per-student federation is meaningful only if grading and storage move to the client; institution-level federation is the alternative for such platforms (Section~\ref{sec:platforms}). Under this threat model, two properties of the exchange matter. First, the global model sent to every sampled client contains the full student embedding table, so each client also receives the embeddings of all other students. Second, secure aggregation hides individual contributions only for parameters that several clients contribute to. A student's row is changed only by that student's client, so the aggregate reveals that student's update exactly. The student embedding is therefore not protected by secure aggregation, which is a further reason to remove it or keep it on the client.

\begin{figure}[!htb]
\centering
\begin{tikzpicture}[font=\scriptsize,
  srv/.style={draw=blue!55!black, rounded corners=3pt, fill=blue!6, minimum width=\linewidth-2pt, minimum height=1.45cm, inner sep=3pt},
  dev/.style={draw=gray!60!black, rounded corners=3pt, fill=white, text width=2.25cm, minimum height=2.6cm, align=center, inner sep=2pt},
  down/.style={-{Stealth[length=5pt]}, line width=1.1pt, blue!60!black},
  up/.style={-{Stealth[length=5pt]}, line width=1.1pt, orange!85!black}]
\node[srv] (S) {};
\node[anchor=west, text=blue!55!black] (sicon) at ([xshift=6pt]S.west) {\fontsize{22}{22}\selectfont\faServer};
\node[anchor=west, align=left, text width=3.4cm] at ([xshift=4pt]sicon.east) {\textbf{Server}\\ combines the students' model updates into one global model (Eq.~\ref{eq:fedavg})};
\node[anchor=east, align=center, text width=2.0cm, text=green!40!black] at ([xshift=-4pt]S.east) {{\large\faShield}\\[1pt] per-skill totals\\ via secure\\ aggregation};
\node[dev] (D1) at ([xshift=1.2cm,yshift=-2.85cm]S.south west) {{\fontsize{20}{20}\selectfont\faLaptop}\\[3pt]
  {\color{blue!55!black}\faGraduationCap}~\textbf{Student 1}\\[4pt]
  {\color{red!60!black}\faLock}~answers \& history\\[2pt]
  {\color{gray!70!black}\faCogs}~trains locally};
\node[dev, right=0.22cm of D1] (D2) {{\fontsize{20}{20}\selectfont\faTablet}\\[3pt]
  {\color{blue!55!black}\faGraduationCap}~\textbf{Student 2}\\[4pt]
  {\color{red!60!black}\faLock}~answers \& history\\[2pt]
  {\color{gray!70!black}\faCogs}~trains locally};
\node[dev, right=0.22cm of D2] (D3) {{\fontsize{20}{20}\selectfont\faMobile}\\[3pt]
  {\color{blue!55!black}\faGraduationCap}~\textbf{Student $K$}\\[4pt]
  {\color{red!60!black}\faLock}~answers \& history\\[2pt]
  {\color{gray!70!black}\faCogs}~trains locally};
\foreach \d in {D1,D2,D3} {
  \draw[down] ([xshift=-6pt]\d.north |- S.south) -- ([xshift=-6pt]\d.north);
  \draw[up]   ([xshift=6pt]\d.north) -- ([xshift=6pt]\d.north |- S.south);
}
\node[anchor=east, text=blue!60!black, align=right] at ([xshift=-8pt,yshift=0.6cm]D1.north) {global\\model};
\node[anchor=west, text=orange!85!black, align=left] at ([xshift=8pt,yshift=0.6cm]D3.north) {model\\update};
\begin{scope}[on background layer]
\node[draw=red!50!black, dashed, rounded corners=4pt, fill=red!3, fit=(D1)(D3), inner sep=5pt] (box) {};
\end{scope}
\node[anchor=north, text=red!50!black] at ([yshift=-1pt]box.south) {{\faLock}~Raw data never leave each student's device};
\end{tikzpicture}
\caption{Data flow of the proposed federated recommender. In each round, the server sends the global model to the sampled students' devices (blue arrows); each device trains it on the student's own answers and returns only the updated model weights (orange arrows), which the server combines into the next global model. Raw interactions, answers, labels, and student features remain on each device (dashed area); aside from the weights, only per-skill totals leave the devices through secure aggregation. Data: own elaboration.}
\label{fig:system}
\end{figure}

\subsection{Hyperparameter optimization}
\label{sec:tuning}

All models are tuned with Optuna \citep{akiba2019optuna} using the multivariate Tree-structured Parzen Estimator (TPE) \citep{bergstra2011tpe}, a Bayesian search method that proposes new configurations close to those that have performed well so far, with grouped conditional parameters to maximize validation AUC. The test set is never loaded during tuning, and tuning uses a random seed (0) disjoint from the evaluation seeds. XGBoost runs 100 trials, and the centralized DNN runs 100 trials with median pruning per epoch, which stops a trial early if its validation AUC falls below the median of earlier trials at that point. Each federated strategy receives a separate study of 30 trials over the local optimizer (SGD or Adam), learning rate, local epochs, local batch size, and weight decay, with median pruning every 50 rounds; the tuned-$\mu$ condition also searches $\mu$. The row-wise condition and the federated logistic regression each receive their own searches over the same space and budget, while the centralized logistic regression receives 100 trials over the optimizer, learning rate, weight decay, and batch size ranges of the centralized DNN. Studies run nine trials in parallel, so the number of started trials slightly exceeds the nominal budget; Table~\ref{tab:search} (Appendix~\ref{app:search}) reports the actual counts. Hyperparameter importance is estimated with fANOVA \citep{hutter2014fanova}, which measures the share of the variation in validation AUC across trials that each hyperparameter explains. Two asymmetries favor the centralized model: the federated architecture is frozen at the centralized optimum, and each federated search receives fewer trials. Both make the measured cost of federation conservative.

\subsection{Evaluation protocol}
\label{sec:evaluation}

Each tuned configuration is retrained with 10 evaluation seeds (42--51). A seed is a number that fixes all random choices of a training run, such as the initial weights and the order of the examples, so that each run can be repeated exactly, and different seeds show how much results vary by chance alone. For federated conditions, a given seed fixes the initial weights, the clients sampled in each round, and the random state of each local update; the conditions are thus compared over identical participant sequences. Test predictions are recorded in each round for trajectory analysis, but the checkpoint and the calibrated threshold are always chosen on validation data.

We report ROC AUC as the primary metric because it is threshold-free and unaffected by the class prior (Section~\ref{sec:measures}). Balanced accuracy is reported at the default threshold of 0.5 and at a \emph{calibrated} threshold that maximizes validation balanced accuracy, that is, Youden's index (Equation~\eqref{eq:balacc}). The F1-Score is reported for comparability, always beside the always-positive reference of Equation~\eqref{eq:f1trivial}, together with the share of pairs predicted positive, which exposes majority-class collapse. Convergence stability is the standard deviation of test AUC across rounds 501--1{,}000 of each run, complemented by its lag-1 autocorrelation, the correlation between the AUC of consecutive rounds, which is close to 1 for a smooth trend and negative for a zigzag oscillation.

The global AUC compares pairs of different students as well as pairs of the same student, and it can therefore be dominated by differences between students. A recommender that chooses the next skill for a given student, however, uses only the ordering of that student's skills. We therefore also report the \emph{per-student AUC}, the AUC computed within each student who has both outcomes in the test window and averaged over these students, and two measures of agreement with the centralized recommender: the median within-student Spearman correlation of the predicted scores, and the share of students for whom both models rank the same skill first.

The simulation runs in PyTorch in an in-process manner, using the aggregation rule in Equation~\eqref{eq:fedavg}. To verify it, we re-ran the tuned FedAvg configuration in the Flower framework \citep{beutel2020flower} with the same client sample and local random state in every round (Section~\ref{sec:flower}).

\subsection{Ablation study}
\label{sec:ablation-design}

The ablation tests the predictions of Section~\ref{sec:embeddings}. First, the tuned centralized DNN and tuned FedAvg are retrained without the student embedding, without the skill embedding, with only the three features, and with only the two embeddings. If federation primarily impairs the embedding tables, the cost of federation should shrink when the tables are removed and grow when the network depends solely on them. Second, the tuned centralized DNN, FedAvg, FedProx $\mu=1.0$ and tuned-$\mu$ configurations are retrained with weight decay restricted to the perceptron, so that no local step shrinks the embedding tables. Third, to observe the tables directly, we record the mean norm of their rows at initialization and after training, as well as the mean distance each row moved from its initial value. Every ablated condition keeps the selected hyperparameters, is not retuned, and is run with the same 10 evaluation seeds, so federated ablations share the client samples from the main runs.

Two robustness analyses complete the design. First, to check that the conclusions do not depend on the random validation split, which shares students and windows with the fitting data, the central conditions are retrained on the first training window and validated on the second, which lies later in each student's history; the test set is unchanged. Second, to separate the proximal term from the local optimizer, FedProx with $\mu=1.0$ and with the tuned $\mu$ is run with the local configuration selected for FedAvg (SGD).

\subsection{Statistical analysis}

Differences between models trained under different paradigms are estimated with Welch's $t$ interval \citep{welch1947}, a confidence interval for the difference of two means that does not assume equal variances, over the ten seeds of each model. Aggregation strategies are compared pairwise with paired $t$-tests over seeds, with Holm correction within each metric \citep{holm1979}, which keeps the chance of a false positive across the ten comparisons at 5\%; we also report the paired effect size $d_z$ (the mean paired difference divided by its standard deviation) and a Friedman omnibus test \citep{demsar2006}, a rank-based test of whether any strategy differs from the others. Equivalence is assessed by two one-sided tests (TOST) with a margin of $\pm 0.01$ AUC \citep{lakens2017equivalence}, which we regard as the smallest difference of practical relevance for ranking recommendations. We do not test differences between strategies over communication rounds of a single run: consecutive rounds are strongly autocorrelated (Section~\ref{sec:stability}), so such tests overstate the evidence. Because the intervals above reflect only the variation between seeds on a fixed test set, the main contrasts are also estimated with a student-cluster bootstrap: each of 2{,}000 replicates resamples the test students with replacement and draws one evaluation seed per model, so that the interval accounts for the sampling of test students as well as for training randomness. Finally, for each pair of strategies, we report the smallest margin at which TOST would still conclude equivalence, ensuring the conclusion does not depend on the choice of $\pm 0.01$.

\section{Results}
\label{sec:results}

\subsection{Protocol check: no leakage signature}

The example construction removes the signature of leakage. For every model, validation and test AUC are close; for the centralized DNN, for example, they are 0.806 and 0.788. The small gap is expected because validation examples come from the same windows as training examples, whereas the test window lies later in each student's history. Under leave-one-out encoding, by contrast, validation exceeded test by 0.21 AUC (Section~\ref{sec:examples}). The models also perform similarly on skills a student has already practiced, and on those new to the student, so they do not depend on having seen the specific pair.

\subsection{Tuning results and reference models}
\label{sec:tuningresults}

Table~\ref{tab:search} (Appendix~\ref{app:search}) lists the search spaces and the selected configurations. For XGBoost, how the student and skill identifiers were encoded mattered most; for the centralized DNN, the regularization settings (weight decay and dropout) mattered most, and the size of the network mattered little.

\begin{figure}[!htb]
\centering
\includegraphics[width=\linewidth]{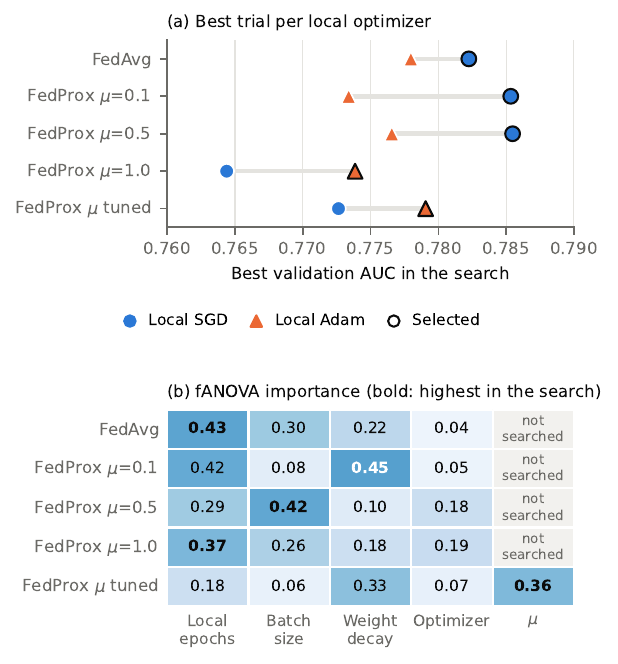}
\caption{Local optimizer in the five federated searches. (a) Best validation AUC reached by trials using local SGD (circles) and local Adam (triangles); the configuration finally selected is outlined in black. (b) fANOVA importance of each searched hyperparameter; the most important one in each search is in bold, and $\mu$ was searched only in the tuned-$\mu$ condition. Data: own experiments.}
\label{fig:optuna}
\end{figure}

In the federated searches, SGD and Adam achieved similar validation AUC across all searches, and the choice of optimizer was never the most important hyperparameter (Figure~\ref{fig:optuna}). What mattered most was how each device trains: the number of local epochs, the batch size, the weight decay, or, in the tuned-$\mu$ search, $\mu$ itself.

\begin{table*}[ht]
\centering
\caption{Test-set performance (mean $\pm$ SD over 10 seeds). Checkpoints and calibrated thresholds are selected on validation data. The first four rows are learning-free references. Data: own experiments.}
\label{tab:main}
\begin{tblr}{
    colspec = {l X[c] X[c] X[c] X[c] X[c]},
    cells   = {font=\small},
    row{1}  = {font=\small\bfseries},
    }
\toprule
Model & ROC AUC & Bal. acc. (0.5) & Bal. acc. (calibr.) & F1 (0.5) & Pred. positive \\
\midrule
Always positive & 0.500 & 0.500 & -- & 0.783 & 1.000 \\
Stratified random guess & 0.503 & 0.503 & -- & 0.658 & 0.669 \\
Student history mean & 0.689 & 0.639 & -- & 0.739 & 0.635 \\
Skill history mean & 0.699 & 0.634 & -- & 0.714 & 0.587 \\
XGBoost, centralized & 0.789 $\pm$ 0.001 & 0.703 $\pm$ 0.003 & 0.706 $\pm$ 0.005 & 0.808 $\pm$ 0.002 & 0.698 $\pm$ 0.007 \\
DNN, centralized & 0.788 $\pm$ 0.002 & 0.690 $\pm$ 0.023 & 0.714 $\pm$ 0.002 & 0.806 $\pm$ 0.013 & 0.716 $\pm$ 0.076 \\
FedAvg & 0.777 $\pm$ 0.001 & 0.670 $\pm$ 0.008 & 0.701 $\pm$ 0.003 & 0.813 $\pm$ 0.003 & 0.774 $\pm$ 0.024 \\
FedProx $\mu=0.1$ & 0.778 $\pm$ 0.001 & 0.615 $\pm$ 0.014 & 0.702 $\pm$ 0.002 & 0.813 $\pm$ 0.002 & 0.881 $\pm$ 0.021 \\
FedProx $\mu=0.5$ & 0.777 $\pm$ 0.001 & 0.672 $\pm$ 0.009 & 0.702 $\pm$ 0.002 & 0.811 $\pm$ 0.003 & 0.764 $\pm$ 0.026 \\
FedProx $\mu=1.0$ & 0.775 $\pm$ 0.003 & 0.500 $\pm$ 0.000 & 0.699 $\pm$ 0.002 & 0.783 $\pm$ 0.000 & 1.000 $\pm$ 0.000 \\
FedProx, $\mu$ tuned & 0.778 $\pm$ 0.004 & 0.500 $\pm$ 0.000 & 0.700 $\pm$ 0.004 & 0.783 $\pm$ 0.000 & 1.000 $\pm$ 0.000 \\
DNN, centralized, preliminary config. & 0.775 $\pm$ 0.003 & 0.684 $\pm$ 0.009 & 0.702 $\pm$ 0.003 & 0.804 $\pm$ 0.004 & 0.722 $\pm$ 0.024 \\
FedAvg, preliminary config. & 0.630 $\pm$ 0.008 & 0.573 $\pm$ 0.015 & 0.600 $\pm$ 0.007 & 0.779 $\pm$ 0.005 & 0.855 $\pm$ 0.039 \\
\bottomrule
\end{tblr}
\end{table*}

\begin{figure*}[ht]
\centering
\includegraphics[width=0.92\textwidth]{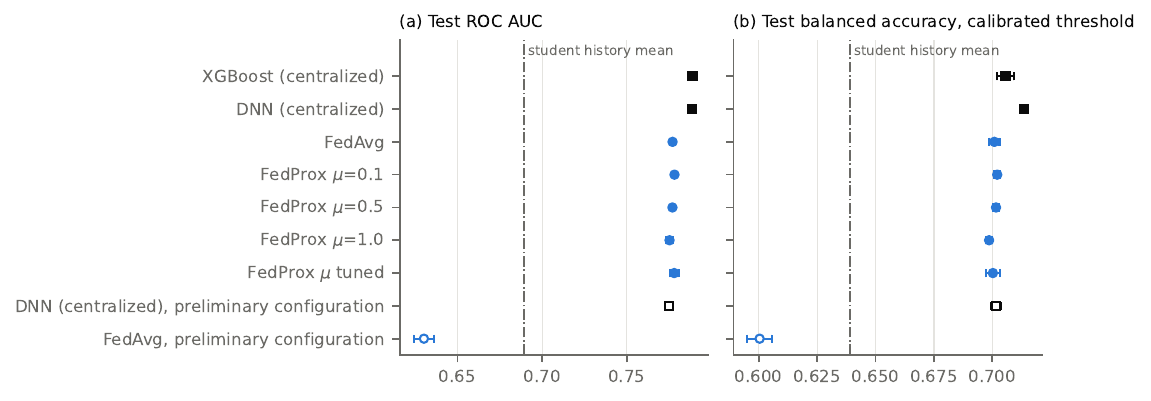}
\caption{Test ROC AUC and calibrated balanced accuracy, mean and 95\% confidence interval over 10 seeds. Squares: centralized models; circles: federated; open markers: configuration of the preliminary analysis. The dash-dotted line marks the learning-free student-history reference. Data: own experiments.}
\label{fig:forest}
\end{figure*}

The two centralized models perform the same (AUC 0.789 for XGBoost and 0.788 for the DNN; Table~\ref{tab:main}), and every tuned model clearly outperforms the learning-free rules, which reach about 0.69 to 0.70 (Figure~\ref{fig:forest}).

\subsection{The cost of federation (RQ1)}

Federation costs about one hundredth of AUC. Tuned FedAvg reaches 0.777, against 0.788 for the centralized DNN of identical architecture, and the FedProx variants lose about the same (Table~\ref{tab:decomposition}, Appendix~\ref{app:supp}). Because the test students are themselves a sample, we also estimated this cost with a bootstrap over students: it remains clearly above zero, but its plausible range is wider, from about 0.004 to 0.019 (Table~\ref{tab:supplementary}). Since the architecture was chosen for centralized training and the federated searches had fewer trials, this is likely a conservative estimate of the cost of federation itself.

\begin{table*}[ht]
\centering
\caption{Global and per-student ROC AUC on the test set (mean $\pm$ SD over 10 seeds), and agreement of federated models with the centralized DNN. Per-student AUC is computed for each of the 825 students with both outcomes in the test window, then averaged; it measures how well a model orders a student's skills. Agreement uses seed-averaged scores: median within-student Spearman correlation, computed over the 949 students with at least three test pairs, and share of students for whom both models rank the same skill first, computed over the 1,063 students with at least two test pairs. Data: own experiments.}
\label{tab:perstudent}
\begin{tblr}{
    colspec = {X[2.6,l] X[1.2,c] X[1.2,c] X[1.1,c] X[1,c]},
    cells   = {font=\small},
    row{1}  = {font=\small\bfseries},
    }
\toprule
Model & Global AUC & Per-student AUC & Spearman with centralized DNN & Same top skill \\
\midrule
Skill history mean & 0.699 & 0.720 & -- & -- \\
XGBoost, centralized & 0.789 $\pm$ 0.001 & 0.731 $\pm$ 0.004 & -- & -- \\
DNN, centralized & 0.788 $\pm$ 0.002 & 0.736 $\pm$ 0.004 & -- & -- \\
Logistic regression, centralized & 0.780 $\pm$ 0.001 & 0.720 & -- & -- \\
FedAvg & 0.777 $\pm$ 0.001 & 0.734 $\pm$ 0.003 & 0.976 & 88.7\% \\
FedProx $\mu=0.1$ & 0.778 $\pm$ 0.001 & 0.732 $\pm$ 0.002 & 0.984 & 92.7\% \\
FedProx $\mu=0.5$ & 0.777 $\pm$ 0.001 & 0.733 $\pm$ 0.002 & 0.986 & 93.0\% \\
FedProx $\mu=1.0$ & 0.775 $\pm$ 0.003 & 0.712 $\pm$ 0.004 & 0.852 & 63.7\% \\
FedProx, $\mu$ tuned & 0.778 $\pm$ 0.004 & 0.712 $\pm$ 0.007 & 0.858 & 69.7\% \\
FedAvg, row-wise (FedAvg config.) & 0.777 $\pm$ 0.001 & 0.738 $\pm$ 0.003 & 0.973 & 84.4\% \\
FedAvg, row-wise (tuned) & 0.767 $\pm$ 0.004 & 0.712 $\pm$ 0.009 & 0.952 & 77.5\% \\
Logistic regression, FedAvg & 0.779 $\pm$ 0.001 & 0.720 & 0.842 & 68.0\% \\
FedAvg, features only & 0.779 $\pm$ 0.000 & 0.720 & 0.842 & 68.0\% \\
FedAvg, no decay on the embedding tables & 0.731 $\pm$ 0.003 & 0.723 $\pm$ 0.006 & -- & -- \\
FedAvg, preliminary config. & 0.630 $\pm$ 0.008 & 0.685 $\pm$ 0.011 & 0.829 & 56.0\% \\
\bottomrule
\end{tblr}
\end{table*}

For recommendation, however, the cost almost disappears. A recommender chooses the next skill for a given student, so what matters is how it orders that student's skills. Measured this way, FedAvg and the centralized DNN are practically equal (per-student AUC 0.734 and 0.736; Table~\ref{tab:perstudent}), and they place the same skill first for 89\% of the students. FedProx with local SGD agrees even more closely with the centralized model, whereas the two configurations with local Adam agree less and rank skills worse. The global cost of federation, therefore, comes mainly from comparing different students, not from ordering each student's skills. For reference, a model that uses only aggregate features ranks each student's skills exactly as the skills' past success rates do (per-student AUC 0.720), because the student's own features are the same across all of that student's skills.

\subsection{Personalization under federation (RQ2)}
\label{sec:ablation}

\begin{table*}[ht]
\centering
\caption{Ablation study: test ROC AUC (mean $\pm$ SD over 10 seeds) of the tuned configurations, not re-tuned. (a) Networks trained with subsets of the inputs; the federation cost is the Welch difference between the centralized DNN and FedAvg. (b) Tuned configurations trained with weight decay on all parameters (``with'', as in the main runs) or on the perceptron only (``without''); the last two columns describe the federated trajectories; values below 0.1\% correspond to the first three rounds, before training takes effect. Data: own experiments.}
\label{tab:ablation}
\begin{tblr}{
    colspec = {X[2.3,l] X[1.3,c] X[1.3,c] X[1.6,c] X[1.5,c]},
    cells   = {font=\small},
    row{2,10}  = {font=\small\bfseries},
    }
\toprule
\SetCell[c=5]{l}\textit{(a) Input blocks} & & & & \\
Inputs & Centralized DNN & FedAvg & Federation cost [95\% CI] & \\
Student emb.\ + skill emb.\ + features (full) & 0.788 $\pm$ 0.002 & 0.777 $\pm$ 0.001 & 0.011 [0.010, 0.013] & \\
Skill emb.\ + features & 0.785 $\pm$ 0.001 & 0.777 $\pm$ 0.002 & 0.008 [0.007, 0.010] & \\
Student emb.\ + features & 0.778 $\pm$ 0.002 & 0.778 $\pm$ 0.001 & 0.000 [$-$0.002, 0.002] & \\
Features only & 0.779 $\pm$ 0.001 & 0.779 $\pm$ 0.000 & 0.000 [0.000, 0.001] & \\
Embeddings only & 0.779 $\pm$ 0.002 & 0.705 $\pm$ 0.002 & 0.074 [0.072, 0.076] & \\
FedAvg, preliminary configuration: full / features only & -- & 0.630 / 0.765 & -- & \\
\midrule
\SetCell[c=5]{l}\textit{(b) Weight decay on the embedding tables} & & & & \\
Condition & AUC, with & AUC, without & SD rounds 501--1{,}000, with / without & Rounds with AUC $<$ 0.5, with / without \\
DNN, centralized & 0.788 $\pm$ 0.002 & 0.780 $\pm$ 0.001 & -- & -- \\
FedAvg & 0.777 $\pm$ 0.001 & 0.731 $\pm$ 0.003 & 0.007 / 0.003 & $<$0.1\% / $<$0.1\% \\
FedProx $\mu=1.0$ & 0.775 $\pm$ 0.003 & 0.658 $\pm$ 0.011 & 0.177 / 0.003 & 40\% / $<$0.1\% \\
FedProx, $\mu$ tuned & 0.778 $\pm$ 0.004 & 0.557 $\pm$ 0.013 & 0.142 / 0.026 & 12\% / 43\% \\
\bottomrule
\end{tblr}
\end{table*}

Removing parts of the model reveals the sources of federation costs (Table~\ref{tab:ablation}), and three findings stand out.

First, the cost lies in the embeddings. Without them, the centralized and federated networks achieve the same AUC (0.779), so federation incurs no additional cost. With them, the centralized network gains about one hundredth of AUC, whereas the federated one gains nothing in global AUC; this difference is the cost of federation. Even in centralized training, most of the gain comes from the skill embedding, and the student embedding adds little. The skill embedding also allows FedAvg to rank each student's skills almost as well as the centralized model; without it, both fall to the level of the features-only model. The student embedding, which is the same across all of a student's skills, can affect this ordering only through interactions within the perceptron, and in practice, it does not.

Second, federated averaging does not learn the student embeddings, as Section~\ref{sec:embeddings} predicted. We measured how much each student's row changed during training (Table~\ref{tab:supplementary}b). Without weight decay on the tables, FedAvg changed the student rows almost not at all in 1{,}000 rounds (by 0.01, for rows of initial size 5.6), whereas centralized training changed them substantially within a few epochs (by 2.5). The tuned configurations address this with weight decay, which shrinks the untrained rows toward zero so they no longer add noise to the input; removing the decay from the tables lowers the AUC of FedAvg from 0.777 to 0.731. The skill rows, which many students update, are also shrunk, but they keep useful information: with decay on the tables, FedAvg ranks each student's skills better than without it and better than without any embeddings (Table~\ref{tab:perstudent}). This comparison, rather than the distance the rows moved, is the evidence, because that distance also includes shrinkage due to decay. Decay also helps the centralized model, so the regularization itself is not specific to federation; what is specific is that, under federation, it is almost the only force acting on the student rows.

Third, making the student embeddings learnable does not help with this data. Row-wise aggregation, the standard remedy in federated recommendation (Equation~\eqref{eq:rowwise}), lets FedAvg train the student rows and raises its validation AUC, but not its test AUC. With the dense FedAvg configuration, test performance remains unchanged; when the method is tuned on its own, test performance is lower (Table~\ref{tab:perstudent}). With a median of nine examples per student, what the model learns about each individual student fits that student's past but does not carry over to later performance, and the validation split, which shares the same windows, rewards it (Section~\ref{sec:limitations}).

Finally, a simple logistic regression on the three aggregate features is as accurate as the recommender in global AUC and loses nothing when federated (0.780 centralized, 0.779 federated). The recommender's only advantage is a better ordering of each student's skills, which it owes to the skill embedding.

\subsection{Aggregation strategies and convergence stability (RQ3)}
\label{sec:stability}

\begin{table*}[ht]
\centering
\caption{Pairwise comparison of the tuned aggregation strategies, paired by seed ($n=10$); $\mu$ values denote FedProx. $\Delta$AUC: test AUC at the validation-selected round; TOST: equivalence test with margin $\pm$0.01 AUC; $\Delta$SD: difference in round-to-round SD of test AUC over rounds 501--1000 (stability). $p$ values Holm-adjusted within each metric. Friedman tests: AUC $\chi^2=13.28$, $p=0.010$; SD $\chi^2=34.16$, $p<0.001$. Data: own experiments.}
\label{tab:strategies}
\begin{tblr}{
    colspec = {X[1.9,l] X[c] X[2,c] X[c] X[0.8,c] X[c] X[c] X[c]},
    cells   = {font=\small},
    row{1}  = {font=\small\bfseries},
    }
\toprule
Pair & $\Delta$AUC & 95\% CI & $p_{\text{Holm}}$ & $d_z$ & TOST $p$ & $\Delta$SD & $p_{\text{Holm}}$ \\
\midrule
FedAvg vs $\mu=0.1$ & $-$0.0011 & [$-$0.0019, $-$0.0004] & 0.055 & $-$1.13 & $<$0.001 & $-$0.0007 & 0.171 \\
FedAvg vs $\mu=0.5$ & 0.0001 & [$-$0.0007, 0.0009] & 1.000 & 0.07 & $<$0.001 & $-$0.0005 & 0.045 \\
FedAvg vs $\mu=1.0$ & 0.0018 & [$-$0.0005, 0.0041] & 0.675 & 0.56 & $<$0.001 & $-$0.1702 & $<$0.001 \\
FedAvg vs $\mu$ tuned & $-$0.0010 & [$-$0.0041, 0.0021] & 1.000 & $-$0.23 & $<$0.001 & $-$0.1347 & $<$0.001 \\
$\mu=0.1$ vs $\mu=0.5$ & 0.0012 & [0.0007, 0.0017] & 0.005 & 1.69 & $<$0.001 & 0.0002 & 0.462 \\
$\mu=0.1$ vs $\mu=1.0$ & 0.0029 & [0.0007, 0.0051] & 0.121 & 0.95 & $<$0.001 & $-$0.1694 & $<$0.001 \\
$\mu=0.1$ vs $\mu$ tuned & 0.0001 & [$-$0.0026, 0.0029] & 1.000 & 0.04 & $<$0.001 & $-$0.1339 & $<$0.001 \\
$\mu=0.5$ vs $\mu=1.0$ & 0.0017 & [$-$0.0005, 0.0039] & 0.675 & 0.55 & $<$0.001 & $-$0.1697 & $<$0.001 \\
$\mu=0.5$ vs $\mu$ tuned & $-$0.0011 & [$-$0.0038, 0.0017] & 1.000 & $-$0.27 & $<$0.001 & $-$0.1342 & $<$0.001 \\
$\mu=1.0$ vs $\mu$ tuned & $-$0.0028 & [$-$0.0051, $-$0.0004] & 0.186 & $-$0.84 & $<$0.001 & 0.0355 & $<$0.001 \\
\bottomrule
\end{tblr}
\end{table*}

At their selected checkpoints, FedAvg and the four FedProx variants are practically interchangeable. They differ by at most 0.003 AUC, and every pair passes the equivalence test, not only with the margin of $\pm$0.01 AUC that we set but also with margins less than half as wide (Table~\ref{tab:strategies}).

\begin{figure*}[ht]
\centering
\includegraphics[width=0.95\textwidth]{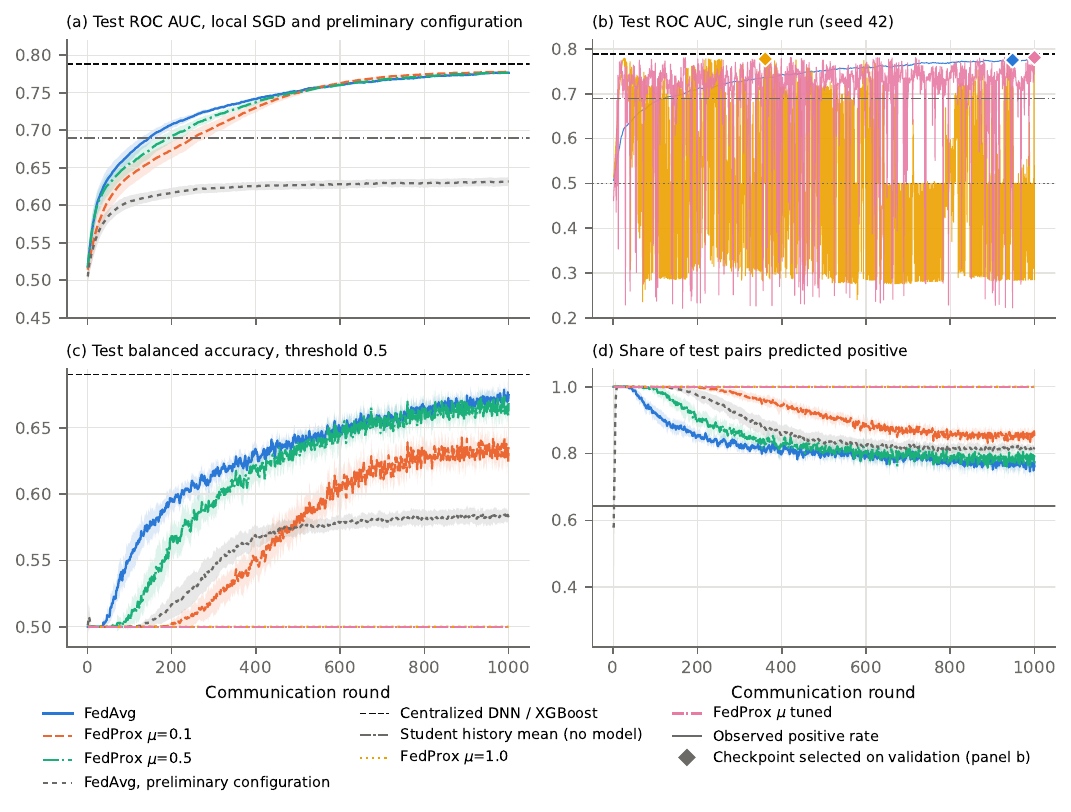}
\caption{Test metrics per communication round. (a), (c) and (d): mean and 95\% confidence band over 10 seeds. (b): a single run (seed 42) of FedAvg and of the two conditions whose tuned configuration uses local Adam with strong weight decay; the mean over seeds would hide their oscillation, and diamonds mark the checkpoint selected on validation data. In (c) and (d), the two Adam conditions lie on 0.5 and 1.0 in every round. The centralized DNN and XGBoost reference lines overlap. Data: own experiments.}
\label{fig:convergence}
\end{figure*}

The strategies differ, however, in how they reach that checkpoint (Figure~\ref{fig:convergence}). The three strategies whose tuning selected local SGD improve smoothly from round to round. The two whose tuning selected local Adam with strong weight decay (FedProx with $\mu=1.0$ and with tuned $\mu$) oscillate from the start: their AUC jumps up and down between rounds, sometimes below chance, and at the default threshold they predict success for virtually every student. Their good AUC at the selected checkpoint comes from a favorable round picked on validation data. For $\mu=1.0$, the ablation identifies the main cause: weight decay acting on the embedding tables through Adam. Without it, the oscillation disappears, although the model then learns more slowly (Table~\ref{tab:ablation}b). For tuned $\mu$, the oscillation persists without the decay, which points to its large learning rate as an additional cause. Because each strategy is compared under its own tuned configuration, the instability belongs to that configuration rather than to the proximal term of FedProx; Section~\ref{sec:robustness} confirms this by running FedProx with the same local optimizer as FedAvg.

\subsection{Robustness of the conclusions}
\label{sec:robustness}

\begin{table*}[ht]
\centering
\caption[Robustness analyses]{Robustness analyses (mean $\pm$ SD over 10 seeds; configurations fixed, not re-tuned). (a) Temporal validation: the models are fitted on the first training window, and the checkpoint is selected on the second, which lies later in every student's history; the test set is unchanged. Because only half of the training examples are used for fitting, absolute values are slightly lower than in the main protocol. (b) FedProx runs with the local configuration selected for FedAvg (SGD), compared with FedAvg and with the Adam configurations selected by the FedProx searches; values below 0.1\% of rounds correspond to the first three rounds, before training takes effect. Data: own experiments.}
\label{tab:robustness}
\begin{tblr}{
    colspec = {X[3,l] X[1.2,c] X[1.2,c] X[1.2,c] X[1.1,c]},
    cells   = {font=\small},
    rowsep  = 1pt,
    row{2,12}  = {font=\small\bfseries},
    }
\toprule
\SetCell[c=5]{l}\textit{(a) Temporal validation: fitted on the first training window, validated on the second} & & & & \\
Condition & Validation AUC & Global test AUC & Per-student AUC & \\
DNN, centralized & 0.790 $\pm$ 0.001 & 0.783 $\pm$ 0.002 & 0.729 $\pm$ 0.007 & \\
DNN, centralized, without student embedding & 0.790 $\pm$ 0.001 & 0.785 $\pm$ 0.001 & 0.733 $\pm$ 0.006 & \\
DNN, centralized, features only & 0.776 $\pm$ 0.000 & 0.778 $\pm$ 0.001 & 0.720 $\pm$ 0.000 & \\
FedAvg & 0.773 $\pm$ 0.002 & 0.767 $\pm$ 0.003 & 0.725 $\pm$ 0.004 & \\
FedAvg, without student embedding & 0.784 $\pm$ 0.002 & 0.777 $\pm$ 0.002 & 0.730 $\pm$ 0.006 & \\
FedAvg, features only & 0.775 $\pm$ 0.000 & 0.778 $\pm$ 0.001 & 0.720 $\pm$ 0.000 & \\
FedAvg, row-wise (FedAvg config.) & 0.761 $\pm$ 0.003 & 0.747 $\pm$ 0.003 & 0.729 $\pm$ 0.005 & \\
FedAvg, row-wise (tuned config.) & 0.750 $\pm$ 0.002 & 0.739 $\pm$ 0.005 & 0.701 $\pm$ 0.009 & \\
\midrule
\SetCell[c=5]{l}\textit{(b) Same local optimizer: FedProx with the SGD configuration of FedAvg} & & & & \\
Condition & Test AUC & SD rounds 501--1{,}000 & Lag-1 autocorrelation & Rounds with AUC $<$ 0.5 \\
FedAvg (SGD) & 0.777 $\pm$ 0.001 & 0.007 & 0.99 & $<$0.1\% \\
FedProx $\mu=1.0$, SGD & 0.769 $\pm$ 0.002 & 0.009 & 0.99 & $<$0.1\% \\
FedProx, $\mu$ tuned, SGD & 0.777 $\pm$ 0.001 & 0.007 & 0.99 & $<$0.1\% \\
FedProx $\mu=1.0$, Adam (tuned config.) & 0.775 $\pm$ 0.003 & 0.177 & $-$0.35 & 40\% \\
FedProx, $\mu$ tuned, Adam (tuned config.) & 0.778 $\pm$ 0.004 & 0.142 & $-$0.06 & 12\% \\
\bottomrule
\end{tblr}
\end{table*}

Two further analyses test whether the conclusions depend on choices of the main protocol (Table~\ref{tab:robustness}).

The first replaces the random validation split with a temporal one: the models are fitted on the first training window, and their checkpoints are chosen on the second, which lies later in each student's history. With this design, validation can no longer reward memorizing students, and the conclusions hold. The cost of federation remains small in global AUC (about 0.016, somewhat larger than before because only half of the training data are used for fitting) and indistinguishable from zero within students, and it again disappears when only the aggregate features are used. The student embedding still adds nothing to the centralized model, and in the federated model, it now hurts: FedAvg without it is about one hundredth of AUC more accurate. Row-wise aggregation, finally, remains worse than dense averaging even when its checkpoint is chosen on the later window, so its failure is not an artifact of the random validation split.

The second analysis addresses the fact that, in the main comparison, FedProx and FedAvg differ not only in the proximal term but also in the local optimizer selected by their searches. Run with the same local configuration as FedAvg (SGD), FedProx is as stable as FedAvg, with no oscillation and no inverted ranking after the first three rounds. With the tuned $\mu$ it matches FedAvg, and with $\mu=1.0$ it loses a little accuracy (0.008 AUC) while remaining within the equivalence margin. The instability reported above, therefore, comes from the local Adam configurations, not from FedProx.

\subsection{Communication budget}

The configurations that use local SGD need several hundred rounds to reach their final quality; within 100 rounds, FedAvg reaches only 0.668 (Table~\ref{tab:budget}, Appendix~\ref{app:supp}). The Adam-based configurations get there sooner but, as shown above, do not stay there. Because all configurations were tuned for 1{,}000 rounds, a search aimed at a shorter horizon could do better.

\subsection{Metrics and the preliminary configuration}

The F1-Score would have told a different story. It ranks FedAvg above both centralized models simply because FedAvg predicts success more often (for 77\% of the test pairs, compared with 64\% observed) and gives the two oscillating strategies the same score as a rule that always predicts success. Calibrating the decision threshold on validation data removes this bias.

The preliminary configuration shows the same problem in a stronger form. It works well when trained centrally, but under federation, it falls below the simplest learning-free rule, while its F1 remains close to that of the always-success rule, making federation look about thirteen times more costly than it is. The mechanism of Section~\ref{sec:ablation} explains why it fails only under federation: it uses no weight decay, so the untrained random embeddings remain in the input as noise. Without the embeddings, the same configuration recovers most of the loss (AUC 0.765 instead of 0.630).

\subsection{Verification against Flower}
\label{sec:flower}

To check our simulator, we re-ran the tuned FedAvg for 100 rounds in the Flower framework \citep{beutel2020flower}, using the same clients and the same random choices in each round. The two implementations agreed to within 0.0001 AUC in every round, so the results above are independent of the simulator. One seed suffices here because, with sampling and randomness matched, additional seeds would exercise the same code.

\section{Discussion}
\label{sec:discussion}

\subsection{Answers to the research questions}

\textbf{RQ1.} Protecting the data costs little. Compared with a centralized recommender with the same architecture, the federated recommender loses about one hundredth of AUC, and this is likely a conservative estimate, since the architecture was chosen for centralized training. The cost is an order of magnitude smaller than the preliminary analysis suggested. For the decision that a recommender actually makes, choosing the next skill for each student, the loss is negligible: the two recommenders order each student's skills almost identically and place the same skill first for 89\% of the students.

\textbf{RQ2.} Only partly, and learning more does not help with these data. Under standard federated averaging, the student embeddings are not learned because each student's contribution is diluted among many others; the skill embeddings, which many students update, are learned well enough to order each student's skills almost as the centralized model does. The local configuration determines how the untrained student embeddings affect the model: without weight decay, they remain noise; with it, they are erased; and under local Adam, the decay caused one configuration to oscillate. Row-wise aggregation, the standard remedy in federated recommendation, trains the student embeddings but does not improve the predictions of later performance. The global cost of federation is essentially what the centralized model gains from its embeddings, and the federated model cannot.

\textbf{RQ3.} No. FedAvg and FedProx reach equivalent predictive performance, and when both use the same local optimizer, they are equally stable (Section~\ref{sec:robustness}); with $\mu=1.0$, FedProx even loses a little accuracy. The instability observed in the main comparison came from the local configurations that the searches selected: those that used local SGD were stable, whereas those that used local Adam with strong weight decay oscillated and, at the default threshold, predicted success for almost every student. This is consistent with Equation~\eqref{eq:dilution}: FedProx averages the returned weights exactly as FedAvg does, so it dilutes the updates of the embedding tables in the same way.

\subsection{Personalization that federation does not learn}

The ablation turns the headline number into a mechanism, and that mechanism directly concerns personalization. Equation~\eqref{eq:dilution} shows why dense averaging cannot train a student's embedding: the row is updated only in the rounds in which that student participates, and then by roughly one hundredth of the local update. The measured movement of 0.01 on a scale of 5.6 confirms that, in practice, the federated recommender never forms a representation of the individual student under dense averaging. It personalizes using the student's aggregate history, which it computes locally, and the skill embedding, which many clients train; together, these two sources are sufficient to rank each student's skills almost as well as the centralized recommender does.

This diagnosis is not new in itself: federated recommendation keeps user parameters on the client precisely because averaging does not train them \citep{ammad2019fcf, singhal2021fedrecon}. What the present study adds is its consequence in student modeling. Removing the dilution with row-wise aggregation allows the federated recommender to learn a representation of each student, but that representation, estimated from a median of nine local examples, fits the student's training windows without improving predictions of the student's later performance. On these data, therefore, the individual student embedding is not where the value of personalization lies; under temporal validation, it even reduces the accuracy of the federated model (Section~\ref{sec:robustness}). Even in centralized training, it contributes only 0.003 AUC, and a federated logistic regression on the aggregate features loses nothing to federation. With more interactions per student, or with institution-level clients, learned student representations could matter more.

A second consequence concerns tuning. The search succeeded for a reason that a practitioner could miss: it found a regularization that erases an untrained component, not a configuration that trains it. Every tuned federated configuration uses weight decay between $3\times10^{-3}$ and $9\times10^{-2}$, in the upper part of the searched range, and the federated searches ranked weight decay, local epochs, and batch size above the choice of optimizer (Section~\ref{sec:tuningresults}). This agrees with guidance that local training must be configured jointly with the aggregation rule \citep{wang2021fieldguide}. Conversely, a validation split that shares students and windows with the training data can reward configurations that memorize students, as the tuned row-wise condition did, so the validation design matters as much as the search.

\subsection{Implications for adaptive learning platforms}
\label{sec:platforms}

From a pedagogical perspective, the question is whether the federated recommender would recommend the same activities as a centralized one. The per-student analysis answers it directly: within each student, the two recommenders order skills almost identically (median Spearman correlation of 0.976) and choose the same top-ranked skill for 89\% of students, with a per-student AUC difference of 0.002, whose interval includes zero. The global cost of 0.011 AUC, which lies at the margin we regarded as practically relevant for comparing strategies, primarily pertains to comparing pairs across different students. It matters for decisions that compare students, such as ranking a class by predicted mastery, more than for the choice of the next activity for each student. Both recommenders, moreover, clearly outperform the simple rules a platform could apply without a model, such as ranking pairs by the student's past success rate (global AUC: 0.689). The decision threshold, however, requires care: at the default threshold of 0.5, the federated recommender would declare 77\% of the pairs mastered, compared with an observed rate of 64\%, and would therefore advance students too early. A platform should calibrate the threshold using validation data or act on skill rankings rather than on a fixed cut-off.

From a data protection perspective, the proposed design maps directly onto the principles of the LGPD. Raw interaction histories, answers, and labels never leave the student's device, which implements the principle of necessity, that is, processing limited to the minimum required for the purpose \citep{LGPD2018}. Student-level features are computed locally, and the skill statistic requires only totals over all students, which secure aggregation can provide without revealing individual contributions \citep{bonawitz2017secagg}. For children and adolescents, whose data must be processed in their best interests, this reduces the data that an institution or platform provider needs to retain and supports the transparency and student control that learning analytics guidelines call for \citep{pardo2014ethical, drachsler2016delicate}.

The results also show that personalization and data protection interact in the model's design. With one client per student and dense averaging, the federated recommender personalizes through each student's aggregate history and the shared skill embedding, but it does not learn a representation of the individual student. The student embedding it carries is the most identifying element of the exchange, since its row changes only when that student participates. A platform has three options. It can remove the student embedding, which, on these data, costs nothing in the federated model and leaves no parameter specific to a single student in the exchange. It can keep the embedding on the student's device and train it locally, as federated collaborative filtering and partially local schemes do \citep{ammad2019fcf, singhal2021fedrecon}; for the student rows, this is computationally equivalent to the row-wise aggregation evaluated here, which trained the embedding but, on these data, did not improve predictions. The equivalence concerns training, not privacy: in the row-wise simulation, the student's row is still sent to the server for aggregation, whereas a locally stored embedding never leaves the device, thereby resolving the privacy weakness identified in Section~\ref{sec:dataflow}. Or it can federate at the level of institutions, where each client holds many students, and the embeddings may become learnable, at the price of centralizing data within each institution.

None of these options provides formal privacy guarantees on its own. Model updates can still reveal information about the data that produced them, so a deployment should combine federation with secure aggregation of the updates, which protects the parameters shared by many clients but not the rows updated by a single one, and, where the risk justifies its cost in accuracy, with differential privacy \citep{abadi2016dpsgd}. Finally, technical measures complement rather than replace the organizational ones: consent, transparency about what the recommender does, and governance of the aggregated model remain the responsibility of the institution \citep{slade2013ethics, oliveira2025fate}.

\subsection{Implications for evaluation practice}

Beyond the research questions, the study shows that evaluation choices can alter conclusions about federated student models, and that six practices follow. Aggregate features should be computed on the history that precedes the example's label window, because leave-one-out encoding leaked the label and produced a validation AUC of 0.21 above the test \citep{prokhorenkova2018catboost}. Threshold-free metrics and learning-free references should accompany any threshold-dependent metric, since the F1-Score ranked the federated model above both centralized models and could not distinguish a collapsed model from the always-positive predictor \citep{pelanek2015metrics}. Both sides of a centralized–federated comparison should share the same architecture and receive comparable tuning, including a local configuration tuned for federation: the preliminary configuration, which used the same architecture on both sides but an untuned local configuration, made the cost of federation appear 13 times larger. Aggregation strategies should be compared across repeated seeds using equivalence tests, and trajectories should be reported, because a checkpoint chosen at the best round obscures whether a configuration can be deployed without a held-out validation set. Federated models with per-client parameters, such as embeddings, should be ablated, and the movement of those parameters should be inspected, because an aggregate metric does not reveal whether they were learned at all. Finally, a model meant to recommend should also be evaluated within students, since a global AUC mixes the ordering of each student's skills with comparisons between students, and here the two told different stories (a cost of 0.011 versus 0.002).

\section{Research Limitations}
\label{sec:limitations}

The findings above are subject to limitations, which we organize following the validity categories of \citet{Wohlin2012}.

\textbf{Internal validity.} Each federated search had 30 nominal trials (38 started, because nine ran in parallel), fewer than the centralized searches, and two selected configurations lie near the upper limit of the weight-decay range; a larger budget or range might find better or more stable configurations for those strategies. Parallel TPE is not exactly reproducible, although the evaluation runs are. The ablated conditions reuse the selected hyperparameters without re-tuning, so they measure each component's contribution to the selected configuration rather than the best performance attainable without it. The input ablation, moreover, covers only FedAvg, and the embedding norms are measured on the final federated model rather than at the selected checkpoint. The validation split is drawn at random from the training windows, so it shares students and windows with the training data. Tuning and early stopping may therefore favor models that memorize students through their embeddings; the larger validation--test gap of the centralized DNN (0.018 AUC) than of FedAvg (0.005) is consistent with this; the temporal-validation analysis (Section~\ref{sec:robustness}) shows that the main conclusions hold when checkpoints are chosen on a later window, although the hyperparameters themselves were tuned on the random split. Finally, the federated simulation is our own implementation; Section~\ref{sec:flower} reports its agreement with Flower for the tuned FedAvg configuration over 100 rounds.

\textbf{Construct validity.} Mastery is operationalized as a success rate of at least 0.70 on a pair within a window, a modeling choice whose alternatives were not evaluated. Our stability measure, the standard deviation of test AUC over the second half of training, also reflects the upward trend of runs that are still improving; for those runs, lag-1 autocorrelation shows that the variation is smooth rather than erratic.

\textbf{External validity.} We deliberately used a single public dataset, because the object of study is the effect of federation on a recommender rather than differences between datasets. The size of the effects may nevertheless differ in other settings: the data were collected in 2009--2010 on a single mathematics tutoring platform in the United States, and Brazilian cohorts, other subjects, and institution-level federations (few clients with many students each) require their own evaluation. Filtering to students with at least 50 interactions excludes the sparsest learners, for whom federated personalization may behave differently.

\textbf{Conclusion validity.} Conclusions rest on ten seeds and a single test split. Seed-only intervals ignore the sampling of test students; the student-cluster bootstrap widens them several-fold (for the cost of federation, from 0.010--0.013 to 0.004--0.019), so differences of a few thousandths of AUC, such as that between FedAvg with and without embeddings, should not be read as exact. The per-student AUC is computed over the 825 students with both outcomes in the test window. The equivalence margin of 0.01 AUC is a judgment about practical relevance; the smallest margins that still yield equivalence (0.0007 to 0.0047) show that the conclusion does not depend on it. Comparisons across paradigms are unpaired because a centralized and a federated run with the same seed do not share their random streams.

\section{Open Science, Reproducibility and Availability of Materials}
\label{sec:openscience}

The complete pipeline, from the raw ASSISTments file (verified by its SHA-256 checksum) to every table and figure in this article, is available at \url{https://github.com/rodrigoronner/federated-recommender-education}. The repository includes the original data file and the examples produced by the pipeline, so that the data used here can be inspected directly. A single script runs the study end-to-end, and every number reported here is read from its output; the tables in Section~\ref{sec:results} are generated by the pipeline itself. Reproducibility rests on three elements. First, all randomness in evaluation derives from the seed: the client sample of each round and the random state of each local update are functions of the seed, the round, and the client, so re-running a condition reproduces its trajectory exactly. Second, software versions are pinned (Python 3.11, PyTorch 2.2, XGBoost 2.0, Optuna 4.1, Flower 1.7). Third, the repository contains the Optuna study journals, the per-round logs of every federated run, the predictions of every evaluated model, and the Flower verification, so that the analyses can be checked without re-training. The ASSISTments data are a public benchmark and contain no directly identifying information. The processed cohort is archived at \url{https://doi.org/10.34740/kaggle/dsv/13081046}, in accordance with the FAIR principles \citep{FAIRprinciples2016} and the UNESCO Recommendation on Open Science \citep{UNESCO2022en}.

\section{Conclusion and Future Work}
\label{sec:conclusion}

This study proposed a federated recommender system for privacy-preserving student performance prediction and asked how much personalization survives federation when each student is a client. On ASSISTments, under a leakage-free protocol with comparable tuning, the federated recommender lost about 0.011 global AUC relative to an identical centralized recommender, and FedAvg and FedProx were equivalent at their selected checkpoints. Within each student, however, the two recommenders ranked skills almost identically, so the choice of the next activity is largely unaffected by federation. The ablation showed what federated training learns. Dense averaging leaves the student embeddings untrained, while the skill embeddings, updated by many clients, carry the within-student ranking. Row-wise aggregation trains the student embeddings, but on these data, this does not improve predictions of later performance, and a federated logistic regression on aggregate features loses nothing to federation. The study also showed how evaluation choices can mislead. Leave-one-out features inflated the validation AUC by 0.21, the F1-Score ranked a federated model above centralized ones, the global AUC overstated the effect of federation on recommendations, and the untuned configuration from the preliminary analysis, which left the random embeddings in its input, made federation appear about 13 times more costly than it is.

The practical message for Informática na Educação is that privacy-preserving personalization is viable at a small cost, provided that the personalized parameters are designed for federation, either kept on the student's device or removed, and that evaluation is designed to detect when federated training fails. For data like this, removing the individual student embedding is the simpler choice: it incurs no cost in the federated model and removes the most identifying parameter from the exchange. Several directions remain open. Locally trained student embeddings \citep{ammad2019fcf, singhal2021fedrecon} should be revisited on data with longer student histories and with a temporal validation window, which would not reward memorizing students. The protocol should be repeated on Brazilian data and on institution-level federations, where each client has many students, and the embeddings may be learnable. Drift-correcting aggregation methods such as SCAFFOLD \citep{karimireddy2020scaffold} and FedDyn \citep{acar2021feddyn} should be compared under the same controls; tuning objectives that reward stable trajectories should replace best-checkpoint selection; and the additional cost of secure aggregation and differential privacy should be measured.

\begin{declarations}

\begin{acknowledgements}
The authors gratefully acknowledge the support of the Software Engineering and Automation Research Laboratory (LaPEA) at the Federal Institute of Education, Science and Technology of Rio Grande do Norte (IFRN), where this research was conducted.
\end{acknowledgements}

\begin{funding}
The authors did not receive support from any organization for the submitted work.
\end{funding}

\begin{contributions}
Conceptualization: R.T. and R.A. Methodology: R.T. Software: R.T. Validation: R.T. and R.A. Formal analysis: R.T. Investigation: R.T. Data curation: R.T. Writing (original draft): R.T. Writing (review and editing): R.T. and R.A. Visualization: R.T. Supervision: R.A. Both authors read and approved the final manuscript.
\end{contributions}

\begin{interests}
The authors declare that they have no conflicts of interest.
\end{interests}

\begin{materials}
The code, the processed examples, the tuning journals and all outputs required to reproduce every number in this article are available at \url{https://github.com/rodrigoronner/federated-recommender-education}. The ASSISTments Skill Builder data is a public, de-identified benchmark distributed by the ASSISTments project; the processed cohort used here is archived at \url{https://doi.org/10.34740/kaggle/dsv/13081046}.
\end{materials}

\begin{aiusagedeclaration}
The authors declare the use of generative AI tools. Claude (Anthropic) was used to review the evaluation protocol, which helped identify the feature leakage and metric artifacts discussed in this article; to refactor the analysis code; and to revise the language of the manuscript. Grammarly was used for language revision. All design decisions and interpretations are the authors' own; every analysis was verified by the authors against the underlying data, and the authors take full responsibility for the content of this article.
\end{aiusagedeclaration}

\begin{furtherinformation}
This study used only publicly available, de-identified secondary data and involved no human participants, no intervention, and no collection of personal data. Approval by a Research Ethics Committee was therefore not applicable under Law No.\ 14,874/2024. Processing complied with the Brazilian General Data Protection Law \citep{LGPD2018}.
\end{furtherinformation}

\end{declarations}

\bibliography{refs}

@article{Baker2011,
    author = {Baker, Ryan and Isotani, Seiji and Carvalho, Adriana},
    title = {{Minera{\c{c}}{\~{a}}o de Dados Educacionais: Oportunidades para o Brasil}},
    doi = {10.5753/rbie.2011.19.02.03},
    issn = {1414-5685},
    journal = {Revista Brasileira de Inform{\'{a}}tica na Educa{\c{c}}{\~{a}}o},
    number = {02},
    volume = {19},
    year = {2011},
    publisher={SBC},
    addendum = "[\href{https://scholar.google.com.br/scholar?q='Mineracao+de+Dados+Educacionais:+Oportunidades+para+o+Brasil'&btnG=Search}{GS Search}]"
}

@ARTICLE{Nascimento2023-cv,
  title     = "{Lei Geral de Prote{\c c}{\~a}o de Dados} ({LGPD}) e
               reposit{\'o}rios institucionais: reflex{\~o}es e adequa{\c
               c}{\~o}es",
  author    = "Nascimento, Bruna and Silva, Edilene",
  journal   = "Em Questão",
  publisher = "FapUNIFESP (SciELO)",
  volume    =  29,
  number    = "e-127314",
  year      =  2023,
 DOI = "10.1590/1808-5245.29.127314"
}

@article{Teodoro2020,
  author    = {Teodoro, Leonardo de Almeida and Kappel, Marco André Abud},
  title     = {Aplicação de Técnicas de Aprendizado de Máquina Para Predição de Risco de Evasão Escolar em Instituições Públicas de Ensino Superior no {Brasil}},
  journal   = {Revista Brasileira de Informática na Educação},
  volume    = {28},
  pages     = {838--863},
  year      = {2020},
  doi       = {10.5753/RBIE.2020.28.0.838}
}

@article{Andrade2025,
  author    = {Andrade, Tiago Luís de and Almeida, Caroline Medeiros Martins de and Barbosa, Jorge Luis Victória and Rigo, Sandro José},
  title     = {Mineração de Dados Educacionais e Metodologias Ativas integrados a um Sistema de Recomendação para prevenção da evasão na Educação a Distância},
  journal   = {Revista Brasileira de Informática na Educação},
  volume    = {33},
  pages     = {748--772},
  year      = {2025},
  doi       = {10.5753/rbie.2025.4641}
}

@incollection{handbook_recsys_2011,
  author    = {Manouselis, Nikos and Drachsler, Hendrik and Verbert, Katrien and Duval, Erik},
  title     = {Recommender Systems in Technology Enhanced Learning},
  booktitle = {Recommender Systems Handbook},
  editor    = {Ricci, Francesco and Rokach, Lior and Shapira, Bracha and Kantor, Paul B.},
  year      = {2011},
  publisher = {Springer US},
  pages     = {387--415},
  isbn      = {978-0-387-85820-3}
}

@article{pereira_deeplearning_2020,
  author    = {Pereira, Filipe Dwan and Fonseca, Samuel C. and Oliveira, Elaine H. T. and Oliveira, David B. F. and Cristea, Alexandra I. and Carvalho, Leandro S. G.},
  title     = {Deep learning for early performance prediction of introductory programming students: a comparative and explanatory study},
  journal   = {Revista Brasileira de Informática na Educação - RBIE},
  volume    = {28},
  pages     = {723--748},
  year      = {2020},
  doi       = {10.5753/RBIE.2020.28.0.723}
}

@article{carneiro_edm_2022,
  author    = {Carneiro, Murillo Guimarães and Dutra, Bruna Luiza and Paiva, José Gustavo S. and Gabriel, Paulo Henrique Ribeiro and Araújo, Rafael Dias},
  title     = {Educational data mining to support identification and prevention of academic retention and dropout: a case study in introductory programming},
  journal   = {Revista Brasileira de Informática na Educação - RBIE},
  volume    = {30},
  pages     = {379--395},
  year      = {2022},
  doi       = {10.5753/rbie.2022.2518}
}

@article{melo_semisupervised_2023,
  author    = {Melo, Cardoso and Souza, Sumika},
  title     = {Improving the prediction of school dropout with the support of the semi-supervised learning approach},
  journal   = {iSys: Revista Brasileira de Sistemas de Informação},
  volume    = {16},
  number    = {1},
  pages     = {10:1--10:26},
  year      = {2023},
  doi       = {10.5753/isys.2023.2852}
}

@article{sales_academic_2016,
  author    = {Sales, Allan and Balby, Leandro and Cajueiro, Adalberto},
  title     = {Exploiting Academic Records for Predicting Student Drop Out: a case study in {Brazilian} higher education},
  journal   = {Journal of Information and Data Management},
  volume    = {7},
  number    = {2},
  pages     = {166--180},
  year      = {2016}
}

@inproceedings{mcmahan_fl_2017,
  author    = {McMahan, H. Brendan and Moore, Eider and Ramage, Daniel and Hampson, Seth and {Ag{\"u}era y Arcas}, Blaise},
  title     = {Communication-Efficient Learning of Deep Networks from Decentralized Data},
  booktitle = {Proceedings of the 20th International Conference on Artificial Intelligence and Statistics (AISTATS)},
  series    = {Proceedings of Machine Learning Research},
  volume    = {54},
  pages     = {1273--1282},
  year      = {2017}
}

@inproceedings{chen_xgboost_2016,
    author    = {Chen, Tianqi and Guestrin, Carlos},
    title     = {{XGBoost}: A Scalable Tree Boosting System},
    booktitle = {Proceedings of the 22nd ACM SIGKDD International Conference on Knowledge Discovery and Data Mining},
    year      = {2016},
    pages     = {785--794},
    publisher = {ACM},
    doi       = {10.1145/2939672.2939785}
}

@article{li_federated_2020,
  author    = {Li, Tian and Sahu, Anit Kumar and Talwalkar, Ameet and Smith, Virginia},
  title     = {Federated Learning: Challenges, Methods, and Future Directions},
  journal   = {IEEE Signal Processing Magazine},
  volume    = {37},
  number    = {3},
  pages     = {50--60},
  year      = {2020},
  doi       = {10.1109/MSP.2020.2975749}
}

@inproceedings{fedprox_2020,
  author    = {Li, Tian and Sahu, Anit Kumar and Zaheer, Manzil and Sanjabi, Maziar and Talwalkar, Ameet and Smith, Virginia},
  title     = {Federated Optimization in Heterogeneous Networks},
  booktitle = {Proceedings of Machine Learning and Systems},
  volume    = {2},
  pages     = {429--450},
  year      = {2020}
}

@inproceedings{hellas_slr_2018,
  author    = {Hellas, Arto and Ihantola, Petri and Petersen, Andrew and Ajanovski, V. V. and Gutica, M. and Hynninen, T. and Knutas, A. and Leinonen, J. and Messom, C. and Liao, S. N.},
  title     = {Predicting academic performance: a systematic literature review},
  booktitle = {Proceedings Companion of the 23rd Annual ACM Conference on Innovation and Technology in Computer Science Education},
  year      = {2018},
  pages     = {175--199},
  doi       = {10.1145/3293881.3295783}
}

@article{colpo_slr_2024,
  author    = {Colpo, Miriam Pizzatto and Thompsen Primo, Tiago and Aguiar, Marilton Sanchotene and Cechinel, Cristian},
  title     = {Educational Data Mining for Dropout Prediction: Trends, Opportunities, and Challenges},
  journal   = {Revista Brasileira de Informática na Educação},
  volume    = {32},
  pages     = {220--256},
  year      = {2024},
  doi       = {10.5753/rbie.2024.3559}
}

@inproceedings{10.1145/3511808.3557108,
author = {Chu, Yun-Wei and Hosseinalipour, Seyyedali and Tenorio, Elizabeth and Cruz, Laura and Douglas, Kerrie and Lan, Andrew and Brinton, Christopher},
title = {Mitigating Biases in Student Performance Prediction via Attention-Based Personalized Federated Learning},
year = {2022},
isbn = {9781450392365},
publisher = {Association for Computing Machinery},
address = {New York, NY, USA},
url = {https://doi.org/10.1145/3511808.3557108},
doi = {10.1145/3511808.3557108},
booktitle = {Proceedings of the 31st ACM International Conference on Information \& Knowledge Management},
pages = {3033–3042},
numpages = {10},
location = {Atlanta, GA, USA},
series = {CIKM '22}
}

@ARTICLE{Feng2009-iu,
  title     = "Addressing the assessment challenge with an online system that
               tutors as it assesses",
  author    = "Feng, Mingyu and Heffernan, Neil and Koedinger, Kenneth",
  journal   = "User Modeling and User-Adapted Interaction",
  publisher = "Springer Science and Business Media LLC",
  volume    =  19,
  number    =  3,
  pages     = "243--266",
  month     =  aug,
  year      =  2009,
  DOI       = "10.1007/s11257-009-9063-7",
  language  = "en"
}

@book{Wohlin2012,
  author    = {Wohlin, Claes and Runeson, Per and H{\"o}st, Martin and Ohlsson, Magnus C. and Regnell, Bj{\"o}rn and Wessl{\'e}n, Anders},
  title     = {Experimentation in Software Engineering},
  publisher = {Springer},
  year      = {2012},
  edition   = {2},
  address   = {Berlin, Heidelberg},
  doi       = {10.1007/978-3-642-29044-2}
}

@misc{LGPD2018,
  author       = {{Brasil}},
  title        = {{Lei n.º 13.709, de 14 de agosto de 2018 — Lei Geral de Proteção de Dados Pessoais (LGPD)}},
  year         = {2018},
  howpublished = {Diário Oficial da União, \url{https://www.planalto.gov.br/ccivil_03/_ato2015-2018/2018/lei/l13709.htm}}
}

@misc{UNESCO2022en,
  author       = {{UNESCO}},
  title        = {{UNESCO} Recommendation on Open Science},
  year         = {2022},
  howpublished = {\url{https://doi.org/10.54677/MNMH8546}},
  doi          = {10.54677/MNMH8546}
}

@article{FAIRprinciples2016,
  author    = {Wilkinson, Mark D. and Dumontier, Michel and Aalbersberg, IJsbrand Jan and Appleton, Gabrielle and Axton, Myles and Baak, Arie and Blomberg, Niklas and Boiten, Jan-Willem and {da Silva Santos}, Luiz Bonino and Bourne, Philip E. and Bouwman, Jildau and Brookes, Anthony J. and Clark, Tim and Crosas, Merc\`{e} and Dillo, Ingrid and Dumon, Olivier and Edmunds, Scott and Evelo, Chris T. and Finkers, Richard and Gonzalez-Beltran, Alejandra and Gray, Alasdair J. G. and Groth, Paul and Goble, Carole and Grethe, Jeffrey S. and Heringa, Jaap and {`t Hoen}, Peter A. C. and Hooft, Rob and Kuhn, Tobias and Kok, Ruben and Kok, Joost and Lusher, Scott J. and Martone, Maryann E. and Mons, Albert and Packer, Abel L. and Persson, Bengt and Rocca-Serra, Philippe and Roos, Marco and {van Schaik}, Rene and Sansone, Susanna-Assunta and Schultes, Erik and Sengstag, Thierry and Slater, Ted and Strawn, George and Swertz, Morris A. and Thompson, Mark and {van der Lei}, Johan and {van Mulligen}, Erik and Velterop, Jan and Waagmeester, Andra and Wittenburg, Peter and Wolstencroft, Katherine and Zhao, Jun and Mons, Barend},
  title     = {The {FAIR} Guiding Principles for Scientific Data Management and Stewardship},
  journal   = {Scientific Data},
  year      = {2016},
  volume    = {3},
  pages     = {160018},
  doi       = {10.1038/sdata.2016.18},
  url       = {https://doi.org/10.1038/sdata.2016.18}
}

@inproceedings{karimireddy2020scaffold,
  author    = {Karimireddy, Sai Praneeth and Kale, Satyen and Mohri, Mehryar and Reddi, Sashank and Stich, Sebastian and Suresh, Ananda Theertha},
  title     = {{SCAFFOLD}: Stochastic Controlled Averaging for Federated Learning},
  booktitle = {Proceedings of the 37th International Conference on Machine Learning},
  series    = {Proceedings of Machine Learning Research},
  volume    = {119},
  pages     = {5132--5143},
  year      = {2020}
}

@inproceedings{acar2021feddyn,
  author    = {Acar, Durmus Alp Emre and Zhao, Yue and Matas, Ramon and Mattina, Matthew and Whatmough, Paul and Saligrama, Venkatesh},
  title     = {Federated Learning Based on Dynamic Regularization},
  booktitle = {International Conference on Learning Representations},
  year      = {2021}
}

@misc{beutel2020flower,
  author        = {Beutel, Daniel J. and Topal, Taner and Mathur, Akhil and Qiu, Xinchi and Fernandez-Marques, Javier and Gao, Yan and Sani, Lorenzo and Li, Kwing Hei and Parcollet, Titouan and de Gusm{\~a}o, Pedro Porto Buarque and Lane, Nicholas D.},
  title         = {Flower: A Friendly Federated Learning Research Framework},
  year          = {2020},
  eprint        = {2007.14390},
  howpublished  = {arXiv preprint arXiv:2007.14390},
  archiveprefix = {arXiv}
}

@inproceedings{akiba2019optuna,
  author    = {Akiba, Takuya and Sano, Shotaro and Yanase, Toshihiko and Ohta, Takeru and Koyama, Masanori},
  title     = {Optuna: A Next-generation Hyperparameter Optimization Framework},
  booktitle = {Proceedings of the 25th ACM SIGKDD International Conference on Knowledge Discovery \& Data Mining},
  pages     = {2623--2631},
  year      = {2019},
  doi       = {10.1145/3292500.3330701}
}

@inproceedings{bergstra2011tpe,
  author    = {Bergstra, James and Bardenet, R{\'e}mi and Bengio, Yoshua and K{\'e}gl, Bal{\'a}zs},
  title     = {Algorithms for Hyper-Parameter Optimization},
  booktitle = {Advances in Neural Information Processing Systems},
  volume    = {24},
  year      = {2011}
}

@inproceedings{piech2015dkt,
  author    = {Piech, Chris and Bassen, Jonathan and Huang, Jonathan and Ganguli, Surya and Sahami, Mehran and Guibas, Leonidas J. and Sohl-Dickstein, Jascha},
  title     = {Deep Knowledge Tracing},
  booktitle = {Advances in Neural Information Processing Systems},
  volume    = {28},
  year      = {2015}
}

@article{corbett1994kt,
  author  = {Corbett, Albert T. and Anderson, John R.},
  title   = {Knowledge Tracing: Modeling the Acquisition of Procedural Knowledge},
  journal = {User Modeling and User-Adapted Interaction},
  volume  = {4},
  number  = {4},
  pages   = {253--278},
  year    = {1994},
  doi     = {10.1007/BF01099821}
}

@article{kaufman2012leakage,
  author  = {Kaufman, Shachar and Rosset, Saharon and Perlich, Claudia and Stitelman, Ori},
  title   = {Leakage in Data Mining: Formulation, Detection, and Avoidance},
  journal = {ACM Transactions on Knowledge Discovery from Data},
  volume  = {6},
  number  = {4},
  pages   = {15:1--15:21},
  year    = {2012},
  doi     = {10.1145/2382577.2382579}
}

@article{kapoor2023leakage,
  author  = {Kapoor, Sayash and Narayanan, Arvind},
  title   = {Leakage and the Reproducibility Crisis in Machine-Learning-Based Science},
  journal = {Patterns},
  volume  = {4},
  number  = {9},
  pages   = {100804},
  year    = {2023},
  doi     = {10.1016/j.patter.2023.100804}
}

@inproceedings{prokhorenkova2018catboost,
  author    = {Prokhorenkova, Liudmila and Gusev, Gleb and Vorobev, Aleksandr and Dorogush, Anna Veronika and Gulin, Andrey},
  title     = {{CatBoost}: Unbiased Boosting with Categorical Features},
  booktitle = {Advances in Neural Information Processing Systems},
  volume    = {31},
  year      = {2018}
}

@article{lakens2017equivalence,
  author  = {Lakens, Dani{\"e}l},
  title   = {Equivalence Tests: A Practical Primer for $t$ Tests, Correlations, and Meta-Analyses},
  journal = {Social Psychological and Personality Science},
  volume  = {8},
  number  = {4},
  pages   = {355--362},
  year    = {2017},
  doi     = {10.1177/1948550617697177}
}

@inproceedings{reddi2021adaptive,
  author    = {Reddi, Sashank J. and Charles, Zachary and Zaheer, Manzil and Garrett, Zachary and Rush, Keith and Kone{\v{c}}n{\'y}, Jakub and Kumar, Sanjiv and McMahan, H. Brendan},
  title     = {Adaptive Federated Optimization},
  booktitle = {International Conference on Learning Representations},
  year      = {2021}
}

@misc{wang2021fieldguide,
  author        = {Wang, Jianyu and Charles, Zachary and Xu, Zheng and Joshi, Gauri and McMahan, H. Brendan and {Aguera y Arcas}, Blaise and Al-Shedivat, Maruan and Andrew, Galen and Avestimehr, Salman and Daly, Katharine and Data, Deepesh and Diggavi, Suhas and Eichner, Hubert and Gadhikar, Advait and Garrett, Zachary and Girgis, Antonious M. and Hanzely, Filip and Hard, Andrew and He, Chaoyang and Horv{\'a}th, Samuel and Huo, Zhouyuan and Ingerman, Alex and Jaggi, Martin and Javidi, Tara and Kairouz, Peter and Kale, Satyen and Karimireddy, Sai Praneeth and Kone{\v{c}}n{\'y}, Jakub and Koyejo, Sanmi and Li, Tian and Liu, Luyang and Mohri, Mehryar and Qi, Hang and Reddi, Sashank J. and Richt{\'a}rik, Peter and Singhal, Karan and Smith, Virginia and Soltanolkotabi, Mahdi and Song, Weikang and Suresh, Ananda Theertha and Stich, Sebastian U. and Talwalkar, Ameet and Wang, Hongyi and Woodworth, Blake and Wu, Shanshan and Yu, Felix X. and Yuan, Honglin and Zaheer, Manzil and Zhang, Mi and Zhang, Tong and Zheng, Chunxiang and Zhu, Chen and Zhu, Wennan},
  title         = {A Field Guide to Federated Optimization},
  year          = {2021},
  eprint        = {2107.06917},
  howpublished  = {arXiv preprint arXiv:2107.06917},
  archiveprefix = {arXiv}
}

@inproceedings{wu2021fdkt,
  author    = {Wu, Jinze and Huang, Zhenya and Liu, Qi and Lian, Defu and Wang, Hao and Chen, Enhong and Ma, Haiping and Wang, Shijin},
  title     = {Federated Deep Knowledge Tracing},
  booktitle = {Proceedings of the 14th ACM International Conference on Web Search and Data Mining},
  pages     = {662--670},
  year      = {2021},
  doi       = {10.1145/3437963.3441747}
}

@article{fachola2023fl,
  author  = {Fachola, Christian and Tornar{\'i}a, Agust{\'i}n and Bermolen, Paola and Capdehourat, Germ{\'a}n and Etcheverry, Lorena and Fariello, Mar{\'i}a In{\'e}s},
  title   = {Federated Learning for Data Analytics in Education},
  journal = {Data},
  volume  = {8},
  number  = {2},
  pages   = {43},
  year    = {2023},
  doi     = {10.3390/data8020043}
}

@article{pelanek2015metrics,
  author  = {Pel{\'a}nek, Radek},
  title   = {Metrics for Evaluation of Student Models},
  journal = {Journal of Educational Data Mining},
  volume  = {7},
  number  = {2},
  pages   = {1--19},
  year    = {2015}
}

@article{demsar2006,
  author  = {Dem{\v{s}}ar, Janez},
  title   = {Statistical Comparisons of Classifiers over Multiple Data Sets},
  journal = {Journal of Machine Learning Research},
  volume  = {7},
  pages   = {1--30},
  year    = {2006}
}

@article{holm1979,
  author  = {Holm, Sture},
  title   = {A Simple Sequentially Rejective Multiple Test Procedure},
  journal = {Scandinavian Journal of Statistics},
  volume  = {6},
  number  = {2},
  pages   = {65--70},
  year    = {1979}
}

@inproceedings{vanhaastrecht2024flame,
  author    = {van Haastrecht, Max and Brinkhuis, Matthieu and Spruit, Marco},
  title     = {Federated Learning Analytics: Investigating the Privacy-Performance Trade-Off in Machine Learning for Educational Analytics},
  booktitle = {Artificial Intelligence in Education (AIED 2024)},
  series    = {Lecture Notes in Computer Science},
  volume    = {14830},
  pages     = {62--74},
  publisher = {Springer},
  year      = {2024},
  doi       = {10.1007/978-3-031-64299-9\_5}
}

@misc{zhao2018noniid,
  author        = {Zhao, Yue and Li, Meng and Lai, Liangzhen and Suda, Naveen and Civin, Damon and Chandra, Vikas},
  title         = {Federated Learning with Non-{IID} Data},
  year          = {2018},
  eprint        = {1806.00582},
  howpublished  = {arXiv preprint arXiv:1806.00582},
  archiveprefix = {arXiv}
}

@article{shemshack2020personalized,
  author  = {Shemshack, Atikah and Spector, Jonathan Michael},
  title   = {A Systematic Literature Review of Personalized Learning Terms},
  journal = {Smart Learning Environments},
  volume  = {7},
  number  = {1},
  pages   = {33},
  year    = {2020},
  doi     = {10.1186/s40561-020-00140-9}
}

@misc{GDPR2016,
  author       = {{European Parliament and Council of the European Union}},
  title        = {Regulation ({EU}) 2016/679 on the Protection of Natural Persons with Regard to the Processing of Personal Data and on the Free Movement of Such Data ({General Data Protection Regulation})},
  year         = {2016},
  howpublished = {Official Journal of the European Union, L 119, pp.~1--88}
}

@inproceedings{bonawitz2017secagg,
  author    = {Bonawitz, Keith and Ivanov, Vladimir and Kreuter, Ben and Marcedone, Antonio and McMahan, H. Brendan and Patel, Sarvar and Ramage, Daniel and Segal, Aaron and Seth, Karn},
  title     = {Practical Secure Aggregation for Privacy-Preserving Machine Learning},
  booktitle = {Proceedings of the 2017 ACM SIGSAC Conference on Computer and Communications Security},
  pages     = {1175--1191},
  year      = {2017},
  doi       = {10.1145/3133956.3133982}
}

@inproceedings{abadi2016dpsgd,
  author    = {Abadi, Mart{\'i}n and Chu, Andy and Goodfellow, Ian and McMahan, H. Brendan and Mironov, Ilya and Talwar, Kunal and Zhang, Li},
  title     = {Deep Learning with Differential Privacy},
  booktitle = {Proceedings of the 2016 ACM SIGSAC Conference on Computer and Communications Security},
  pages     = {308--318},
  year      = {2016},
  doi       = {10.1145/2976749.2978318}
}

@inproceedings{singhal2021fedrecon,
  author    = {Singhal, Karan and Sidahmed, Hakim and Garrett, Zachary and Wu, Shanshan and Rush, John and Prakash, Sushant},
  title     = {Federated Reconstruction: Partially Local Federated Learning},
  booktitle = {Advances in Neural Information Processing Systems},
  volume    = {34},
  year      = {2021}
}

@inproceedings{xiong2016deeper,
  author    = {Xiong, Xiaolu and Zhao, Siyuan and Van Inwegen, Eric G. and Beck, Joseph E.},
  title     = {Going Deeper with Deep Knowledge Tracing},
  booktitle = {Proceedings of the 9th International Conference on Educational Data Mining},
  pages     = {545--550},
  year      = {2016}
}

@article{gervet2020kt,
  author  = {Gervet, Theophile and Koedinger, Ken and Schneider, Jeff and Mitchell, Tom},
  title   = {When Is Deep Learning the Best Approach to Knowledge Tracing?},
  journal = {Journal of Educational Data Mining},
  volume  = {12},
  number  = {3},
  pages   = {31--54},
  year    = {2020},
  doi     = {10.5281/zenodo.4143614}
}

@article{sarsa2022dlkt,
  author  = {Sarsa, Sami and Leinonen, Juho and Hellas, Arto},
  title   = {Empirical Evaluation of Deep Learning Models for Knowledge Tracing: Of Hyperparameters and Metrics on Performance and Replicability},
  journal = {Journal of Educational Data Mining},
  volume  = {14},
  number  = {2},
  pages   = {32--102},
  year    = {2022},
  doi     = {10.5281/zenodo.7086179}
}

@inproceedings{hutter2014fanova,
  author    = {Hutter, Frank and Hoos, Holger and Leyton-Brown, Kevin},
  title     = {An Efficient Approach for Assessing Hyperparameter Importance},
  booktitle = {Proceedings of the 31st International Conference on Machine Learning},
  series    = {Proceedings of Machine Learning Research},
  volume    = {32},
  pages     = {754--762},
  year      = {2014}
}

@inproceedings{loshchilov2019adamw,
  author    = {Loshchilov, Ilya and Hutter, Frank},
  title     = {Decoupled Weight Decay Regularization},
  booktitle = {International Conference on Learning Representations},
  year      = {2019}
}

@article{youden1950,
  author  = {Youden, W. J.},
  title   = {Index for Rating Diagnostic Tests},
  journal = {Cancer},
  volume  = {3},
  number  = {1},
  pages   = {32--35},
  year    = {1950},
  doi     = {10.1002/1097-0142(1950)3:1<32::AID-CNCR2820030106>3.0.CO;2-3}
}

@article{fawcett2006roc,
  author  = {Fawcett, Tom},
  title   = {An Introduction to {ROC} Analysis},
  journal = {Pattern Recognition Letters},
  volume  = {27},
  number  = {8},
  pages   = {861--874},
  year    = {2006},
  doi     = {10.1016/j.patrec.2005.10.010}
}

@inproceedings{brodersen2010balanced,
  author    = {Brodersen, Kay Henning and Ong, Cheng Soon and Stephan, Klaas Enno and Buhmann, Joachim M.},
  title     = {The Balanced Accuracy and Its Posterior Distribution},
  booktitle = {Proceedings of the 20th International Conference on Pattern Recognition},
  pages     = {3121--3124},
  year      = {2010},
  doi       = {10.1109/ICPR.2010.764}
}

@article{welch1947,
  author  = {Welch, B. L.},
  title   = {The Generalization of `{Student's}' Problem when Several Different Population Variances Are Involved},
  journal = {Biometrika},
  volume  = {34},
  number  = {1--2},
  pages   = {28--35},
  year    = {1947},
  doi     = {10.1093/biomet/34.1-2.28}
}

@article{bloom1984sigma,
  author  = {Bloom, Benjamin S.},
  title   = {The 2 Sigma Problem: The Search for Methods of Group Instruction as Effective as One-to-One Tutoring},
  journal = {Educational Researcher},
  volume  = {13},
  number  = {6},
  pages   = {4--16},
  year    = {1984},
  doi     = {10.3102/0013189X013006004}
}

@article{vanlehn2011tutoring,
  author  = {VanLehn, Kurt},
  title   = {The Relative Effectiveness of Human Tutoring, Intelligent Tutoring Systems, and Other Tutoring Systems},
  journal = {Educational Psychologist},
  volume  = {46},
  number  = {4},
  pages   = {197--221},
  year    = {2011},
  doi     = {10.1080/00461520.2011.611369}
}

@incollection{drachsler2015panorama,
  author    = {Drachsler, Hendrik and Verbert, Katrien and Santos, Olga C. and Manouselis, Nikos},
  title     = {Panorama of Recommender Systems to Support Learning},
  booktitle = {Recommender Systems Handbook},
  editor    = {Ricci, Francesco and Rokach, Lior and Shapira, Bracha},
  edition   = {2},
  publisher = {Springer},
  address   = {Boston, MA},
  pages     = {421--451},
  year      = {2015},
  doi       = {10.1007/978-1-4899-7637-6\_12}
}

@article{slade2013ethics,
  author  = {Slade, Sharon and Prinsloo, Paul},
  title   = {Learning Analytics: Ethical Issues and Dilemmas},
  journal = {American Behavioral Scientist},
  volume  = {57},
  number  = {10},
  pages   = {1510--1529},
  year    = {2013},
  doi     = {10.1177/0002764213479366}
}

@article{pardo2014ethical,
  author  = {Pardo, Abelardo and Siemens, George},
  title   = {Ethical and Privacy Principles for Learning Analytics},
  journal = {British Journal of Educational Technology},
  volume  = {45},
  number  = {3},
  pages   = {438--450},
  year    = {2014},
  doi     = {10.1111/bjet.12152}
}

@inproceedings{drachsler2016delicate,
  author    = {Drachsler, Hendrik and Greller, Wolfgang},
  title     = {Privacy and Analytics: It's a {DELICATE} Issue. A Checklist for Trusted Learning Analytics},
  booktitle = {Proceedings of the Sixth International Conference on Learning Analytics \& Knowledge (LAK '16)},
  pages     = {89--98},
  publisher = {ACM},
  year      = {2016},
  doi       = {10.1145/2883851.2883893}
}

@article{ifenthaler2016perceptions,
  author  = {Ifenthaler, Dirk and Schumacher, Clara},
  title   = {Student Perceptions of Privacy Principles for Learning Analytics},
  journal = {Educational Technology Research and Development},
  volume  = {64},
  number  = {5},
  pages   = {923--938},
  year    = {2016},
  doi     = {10.1007/s11423-016-9477-y}
}

@article{oliveira2025fate,
  author  = {Oliveira, Letícia Rocha de and Matos, Diego Dermeval Medeiros da Cunha and Rodrigues, Luiz Antônio Lima and Lima, Maria Luiza Correia Oliveira and Lima, Bárbara de},
  title   = {Justiça, Responsabilidade, Transparência e Ética em Learning Analytics: Um Mapeamento Sistemático da Literatura},
  journal = {Revista Brasileira de Informática na Educação},
  volume  = {33},
  pages   = {773--798},
  year    = {2025},
  doi     = {10.5753/rbie.2025.5840}
}

@inproceedings{maretto2023dados,
  author    = {Maretto, Ciro X. and Barcellos, Monalessa P.},
  title     = {Uma Avaliação de Dados Abertos Educacionais Brasileiros: Qualidade, Privacidade e {Learning Analytics}},
  booktitle = {Anais do XXXIV Simpósio Brasileiro de Informática na Educação (SBIE 2023)},
  pages     = {1341--1350},
  publisher = {SBC},
  year      = {2023},
  doi       = {10.5753/sbie.2023.234758}
}

@inproceedings{dias2020recommender,
  author    = {Dias, Alessandro da Silveira and Wives, Leandro Krug},
  title     = {Recommender System for Learning Objects Based in the Results of Learner Choices in E-learning Systems: Application in the {MERLOT} Platform},
  booktitle = {Anais do XXXI Simpósio Brasileiro de Informática na Educação (SBIE 2020)},
  pages     = {922--931},
  publisher = {SBC},
  year      = {2020},
  doi       = {10.5753/cbie.sbie.2020.922}
}

@misc{ammad2019fcf,
  author        = {Ammad-ud-din, Muhammad and Ivannikova, Elena and Khan, Suleiman A. and Oyomno, Were and Fu, Qiang and Tan, Kuan Eeik and Flanagan, Adrian},
  title         = {Federated Collaborative Filtering for Privacy-Preserving Personalized Recommendation System},
  year          = {2019},
  eprint        = {1901.09888},
  howpublished  = {arXiv preprint arXiv:1901.09888},
  archiveprefix = {arXiv}
}

@article{chai2021sfmf,
  author  = {Chai, Di and Wang, Leye and Chen, Kai and Yang, Qiang},
  title   = {Secure Federated Matrix Factorization},
  journal = {IEEE Intelligent Systems},
  volume  = {36},
  number  = {5},
  pages   = {11--20},
  year    = {2021},
  doi     = {10.1109/MIS.2020.3014880}
}

@article{lin2021fedrec,
  author  = {Lin, Guanyu and Liang, Feng and Pan, Weike and Ming, Zhong},
  title   = {{FedRec}: Federated Recommendation With Explicit Feedback},
  journal = {IEEE Intelligent Systems},
  volume  = {36},
  number  = {5},
  pages   = {21--30},
  year    = {2021},
  doi     = {10.1109/MIS.2020.3017205}
}

@inproceedings{muhammad2020fedfast,
  author    = {Muhammad, Khalil and Wang, Qinqin and O'Reilly-Morgan, Diarmuid and Tragos, Elias and Smyth, Barry and Hurley, Neil and Geraci, James and Lawlor, Aonghus},
  title     = {{FedFast}: Going Beyond Average for Faster Training of Federated Recommender Systems},
  booktitle = {Proceedings of the 26th ACM SIGKDD International Conference on Knowledge Discovery \& Data Mining},
  pages     = {1234--1242},
  year      = {2020},
  doi       = {10.1145/3394486.3403176}
}

@misc{arivazhagan2019fedper,
  author        = {Arivazhagan, Manoj Ghuhan and Aggarwal, Vinay and Singh, Aaditya Kumar and Choudhary, Sunav},
  title         = {Federated Learning with Personalization Layers},
  year          = {2019},
  eprint        = {1912.00818},
  howpublished  = {arXiv preprint arXiv:1912.00818},
  archiveprefix = {arXiv}
}

@misc{anonymous2025preliminary,
  author       = {{Anonymous}},
  title        = {Preliminary analysis of federated recommendation for student performance prediction},
  year         = {2025},
  howpublished = {Details omitted for double-blind review}
}

\clearpage
\setlength{\intextsep}{8pt}
\setlength{\floatsep}{8pt}
\setlength{\textfloatsep}{8pt}
\appendix
\section{Hyperparameter search}
\label{app:search}

\begin{table*}[!htbp]
\centering
\caption{Optuna search spaces (TPE sampler, validation AUC objective) and selected configurations. Federated space: local optimizer \{SGD, Adam\} with the learning-rate ranges of the centralized search (Adam up to $10^{-1}$), local epochs $E\in\{1,2,5,10\}$, batch size $B\in\{16,32,64\}$, weight decay $10^{-7}$--$10^{-1}$, and $\mu\in[10^{-3},1]$ (log) for the tuned-$\mu$ condition. Data: own experiments.}
\label{tab:search}
\begin{tblr}{
    colspec = {p{0.40\textwidth} p{0.31\textwidth} X},
    cells   = {font=\small},
    rowsep  = 1pt,
    row{1}  = {font=\small\bfseries},
    }
\toprule
Hyperparameter & Search space & Selected \\
\midrule
\SetCell[c=3]{l}\textit{XGBoost} (108 trials) &  &  \\
ID encoding & numeric, categorical, none & numeric \\
Max. depth; learning rate & 2--10; $10^{-2}$--0.3 (log) & 6; 0.060 \\
Min. child weight; subsample; col. sample & 1--50 (log); 0.5--1; 0.5--1 & 3.49; 0.68; 0.58 \\
$\lambda$; $\alpha$; $\gamma$ & $10^{-3}$--10 (log); $10^{-3}$--10 (log); 0--5 & 0.026; 0.067; 1.07 \\
\SetCell[c=3]{l}\textit{RecommenderNet, centralized} (108 trials) &  &  \\
Embedding; hidden 1; hidden 2 & \{2,4,8,16,32\}; \{16,\ldots,256\}; \{8,\ldots,128\} & 32; 128; 128 \\
Dropout; weight decay & 0--0.7; $10^{-7}$--$10^{-1}$ (log) & 0.4; $3.1\times10^{-3}$ \\
Optimizer; learning rate & Adam $10^{-4}$--$3\cdot10^{-2}$, SGD $10^{-2}$--2 (log) & Adam; 0.0055 \\
Batch size & \{32,64,128,256\} & 64 \\
\SetCell[c=3]{l}\textit{Federated training} (38 trials per strategy; architecture frozen from the centralized search) &  &  \\
FedAvg & \SetCell[c=2]{l}SGD, lr 0.2171, $E$=2, $B$=32, weight decay $3.1\times10^{-3}$ &  \\
FedProx $\mu=0.1$ & \SetCell[c=2]{l}SGD, lr 0.1459, $E$=1, $B$=16, weight decay $8.7\times10^{-3}$ &  \\
FedProx $\mu=0.5$ & \SetCell[c=2]{l}SGD, lr 0.3609, $E$=1, $B$=16, weight decay $3.1\times10^{-3}$ &  \\
FedProx $\mu=1.0$ & \SetCell[c=2]{l}Adam, lr 0.0137, $E$=10, $B$=16, weight decay $7.9\times10^{-2}$ &  \\
FedProx, $\mu$ tuned & \SetCell[c=2]{l}Adam, lr 0.0713, $E$=1, $B$=16, weight decay $9.0\times10^{-2}$; $\mu$=0.017 &  \\
\SetCell[c=3]{l}\textit{Added conditions} (search spaces as above) &  &  \\
FedAvg, row-wise aggregation (38 trials) & \SetCell[c=2]{l}SGD, lr 0.3043, $E$=5, $B$=16, weight decay $1.8\times10^{-3}$ &  \\
Logistic regression, FedAvg (38 trials) & \SetCell[c=2]{l}SGD, lr 0.0915, $E$=5, $B$=16, weight decay $2.0\times10^{-7}$ &  \\
Logistic regression, centralized (108 trials) & \SetCell[c=2]{l}SGD, lr 1.8097, $B$=128, weight decay $3.6\times10^{-7}$ &  \\
\bottomrule
\end{tblr}
\end{table*}

\section{Supplementary results}
\label{app:supp}

\begin{table*}[!htbp]
\centering
\caption[Architecture effect and cost of federation]{Architecture effect and cost of federation: difference of means with Welch 95\% confidence interval over 10 seeds per model, and Cohen's $d$. Because $d$ divides the difference by the variation between seeds, which is very small here, large values of $d$ indicate that a difference is consistent across repetitions, not that it is large in practice; its practical size is the difference in AUC or balanced accuracy itself. Data: own experiments.}
\label{tab:decomposition}
\begin{tblr}{
    colspec = {X[2.6,l] X[1.5,c] X[0.5,c] X[1.5,c] X[0.5,c]},
    cells   = {font=\small},
    rowsep  = 1pt,
    row{1}  = {font=\small\bfseries},
    }
\toprule
Contrast & ROC AUC & $d$ & Balanced accuracy (calibrated) & $d$ \\
\midrule
Architecture: XGBoost $-$ centralized DNN & 0.000 [$-$0.001, 0.002] & 0.2 & $-$0.008 [$-$0.012, $-$0.004] & $-$2.1 \\
Federation cost: centralized DNN $-$ FedAvg & 0.011 [0.010, 0.013] & 7.9 & 0.013 [0.010, 0.015] & 4.9 \\
Federation cost: centralized DNN $-$ FedProx $\mu=0.1$ & 0.010 [0.009, 0.012] & 8.2 & 0.012 [0.010, 0.013] & 5.8 \\
Federation cost: centralized DNN $-$ FedProx $\mu=0.5$ & 0.012 [0.010, 0.013] & 8.1 & 0.012 [0.010, 0.014] & 5.8 \\
Federation cost: centralized DNN $-$ FedProx $\mu=1.0$ & 0.013 [0.011, 0.016] & 5.6 & 0.015 [0.013, 0.017] & 7.4 \\
Federation cost: centralized DNN $-$ FedProx, $\mu$ tuned & 0.011 [0.008, 0.013] & 3.6 & 0.013 [0.010, 0.017] & 4.0 \\
Federation cost, preliminary configuration & 0.145 [0.139, 0.151] & 22.7 & 0.101 [0.096, 0.107] & 18.2 \\
FedAvg $-$ centralized XGBoost & $-$0.012 [$-$0.013, $-$0.011] & $-$10.1 & $-$0.005 [$-$0.009, $-$0.001] & $-$1.2 \\
\bottomrule
\end{tblr}
\end{table*}

\begin{table*}[!htbp]
\centering
\caption{Test ROC AUC (mean $\pm$ SD over 10 seeds) when the checkpoint is chosen on validation data within a budget of $R$ communication rounds. Configurations were tuned for $R=1{,}000$. Data: own experiments.}
\label{tab:budget}
\begin{tblr}{
    colspec = {l X[c] X[c] X[c] X[c]},
    cells   = {font=\small},
    rowsep  = 1pt,
    row{1}  = {font=\small\bfseries},
    }
\toprule
Condition & $R=100$ & $R=250$ & $R=500$ & $R=1{,}000$ \\
\midrule
FedAvg & 0.668 $\pm$ 0.015 & 0.718 $\pm$ 0.003 & 0.752 $\pm$ 0.003 & 0.777 $\pm$ 0.001 \\
FedProx $\mu=0.1$ & 0.639 $\pm$ 0.019 & 0.689 $\pm$ 0.014 & 0.750 $\pm$ 0.005 & 0.778 $\pm$ 0.001 \\
FedProx $\mu=0.5$ & 0.655 $\pm$ 0.016 & 0.706 $\pm$ 0.008 & 0.751 $\pm$ 0.003 & 0.777 $\pm$ 0.001 \\
FedProx $\mu=1.0$ & 0.773 $\pm$ 0.007 & 0.774 $\pm$ 0.006 & 0.775 $\pm$ 0.003 & 0.775 $\pm$ 0.003 \\
FedProx, $\mu$ tuned & 0.777 $\pm$ 0.002 & 0.778 $\pm$ 0.001 & 0.778 $\pm$ 0.003 & 0.778 $\pm$ 0.004 \\
FedAvg, preliminary config. & 0.604 $\pm$ 0.011 & 0.619 $\pm$ 0.011 & 0.625 $\pm$ 0.011 & 0.630 $\pm$ 0.008 \\
\bottomrule
\end{tblr}
\end{table*}

\begin{table*}[!htbp]
\centering
\caption[Supplementary estimates]{Supplementary estimates. (a) Differences in test AUC with 95\% confidence intervals from a student-cluster bootstrap (2{,}000 replicates, one random evaluation seed per model in each replicate), which account for the sampling of test students as well as for training randomness. (b) Mean norm of the embedding rows at the end of training and mean distance each row moved from its initial value; federated models are taken at round 1{,}000, centralized models at the selected epoch. The distance moved includes the shrinkage caused by weight decay and therefore does not by itself indicate learning. Data: own experiments.}
\label{tab:supplementary}
\begin{tblr}{
    colspec = {X[3.4,l] X[1,c] X[1,c] X[1,c] X[1,c]},
    cells   = {font=\small},
    rowsep  = 1pt,
    row{2,17}  = {font=\small\bfseries},
    }
\toprule
\SetCell[c=5]{l}\textit{(a) Contrasts with student-cluster bootstrap 95\% CI} & & & & \\
Contrast & \SetCell[c=2]{c}Global AUC & & \SetCell[c=2]{c}Per-student AUC & \\
Cost of federation: DNN, centralized $-$ FedAvg & \SetCell[c=2]{c}0.011 [0.004, 0.019] & & \SetCell[c=2]{c}0.002 [$-$0.011, 0.015] & \\
Cost of federation: DNN, centralized $-$ FedProx $\mu=0.1$ & \SetCell[c=2]{c}0.010 [0.004, 0.016] & & \SetCell[c=2]{c}0.004 [$-$0.008, 0.016] & \\
Cost of federation: DNN, centralized $-$ FedProx $\mu=0.5$ & \SetCell[c=2]{c}0.012 [0.005, 0.018] & & \SetCell[c=2]{c}0.003 [$-$0.010, 0.015] & \\
Cost of federation: DNN, centralized $-$ FedProx $\mu=1.0$ & \SetCell[c=2]{c}0.013 [0.006, 0.022] & & \SetCell[c=2]{c}0.024 [0.007, 0.042] & \\
Cost of federation: DNN, centralized $-$ FedProx, $\mu$ tuned & \SetCell[c=2]{c}0.011 [0.002, 0.022] & & \SetCell[c=2]{c}0.024 [0.005, 0.048] & \\
Cost of federation, features only & \SetCell[c=2]{c}0.000 [$-$0.003, 0.003] & & \SetCell[c=2]{c}0.000 [0.000, 0.000] & \\
Gain from embeddings, centralized & \SetCell[c=2]{c}0.009 [0.002, 0.015] & & \SetCell[c=2]{c}0.016 [0.000, 0.031] & \\
FedAvg: features only $-$ full network & \SetCell[c=2]{c}0.002 [$-$0.005, 0.009] & & \SetCell[c=2]{c}$-$0.014 [$-$0.030, 0.003] & \\
FedAvg: row-wise (tuned) $-$ dense & \SetCell[c=2]{c}$-$0.010 [$-$0.022, 0.000] & & \SetCell[c=2]{c}$-$0.023 [$-$0.045, $-$0.003] & \\
Cost of federation, logistic regression & \SetCell[c=2]{c}0.000 [$-$0.003, 0.004] & & \SetCell[c=2]{c}0.000 [0.000, 0.000] & \\
DNN $-$ logistic regression, centralized & \SetCell[c=2]{c}0.009 [0.002, 0.016] & & \SetCell[c=2]{c}0.016 [0.000, 0.032] & \\
FedAvg: recommender $-$ logistic regression & \SetCell[c=2]{c}$-$0.002 [$-$0.010, 0.006] & & \SetCell[c=2]{c}0.014 [$-$0.002, 0.030] & \\
Cost of federation, preliminary configuration & \SetCell[c=2]{c}0.145 [0.123, 0.167] & & \SetCell[c=2]{c}0.033 [0.004, 0.069] & \\
\midrule
\SetCell[c=5]{l}\textit{(b) Embedding rows after training (mean over 10 seeds; initial mean norm 5.6)} & & & & \\
Condition & Student rows: norm & Student rows: moved & Skill rows: norm & Skill rows: moved \\
DNN, centralized, decay on all parameters & 0.09 & 5.61 & 0.18 & 5.62 \\
DNN, centralized, no decay on the tables & 6.09 & 2.55 & 5.79 & 2.53 \\
FedAvg, decay on all parameters & 0.76 & 4.85 & 0.81 & 4.82 \\
FedAvg, no decay on the tables & 5.61 & 0.01 & 5.65 & 0.12 \\
FedAvg, row-wise (FedAvg config.) & 4.74 & 1.79 & 0.92 & 4.87 \\
FedAvg, row-wise (tuned) & 2.20 & 4.42 & 0.46 & 5.49 \\
\bottomrule
\end{tblr}
\end{table*}

\end{document}